\documentclass[authoryear,preprint,12pt]{elsarticle}

\usepackage{amsmath,amssymb,amsfonts}
\usepackage{graphicx}
\usepackage{textcomp}
\usepackage{xcolor}
\usepackage{tabularx}
\usepackage{subcaption}
\usepackage[english]{babel}
\usepackage{amsthm}
\usepackage{amsmath}
\usepackage{algorithm}
\usepackage{algorithmic}
\usepackage{graphicx}
\usepackage{hyperref}
\usepackage{xurl}
\usepackage[inline]{enumitem}

\DeclareUnicodeCharacter{2212}{-}

\theoremstyle{definition}
\newtheorem{definition}{Definition}

\theoremstyle{remark}

\def\BibTeX{{\rm B\kern-.05em{\sc i\kern-.025em b}\kern-.08em
    T\kern-.1667em\lower.7ex\hbox{E}\kern-.125emX}}

\journal{Expert Systems with Applications}

\begin{document}

\begin{frontmatter}



\title{Energy-Predictive Planning for Optimizing Drone Service Delivery}


\author[1]{Guanting Ren}
\ead{gren5848@uni.sydney.edu.au}

\author[2]{Babar Shahzaad\corref{cor1}}
\ead{babar.shahzaad@qut.edu.au}

\author[1]{Balsam Alkouz}
\ead{balsam.alkouz@sydney.edu.au}

\author[1]{Abdallah Lakhdari}
\ead{abdallah.lakhdari@sydney.edu.au}

\author[1]{Athman Bouguettaya}
\ead{athman.bouguettaya@sydney.edu.au}

\affiliation[1]{organization={School of Computer Science, The University of Sydney},
            addressline={}, 
            city={Sydney},
            postcode={2006}, 
            state={NSW},
            country={Australia}}

\affiliation[2]{organization={Queensland University of Technology},
    city={Brisbane},
    postcode={4000}, 
    state={QLD},
    country={Australia}}

\cortext[cor1]{Corresponding author}

\begin{abstract}
We propose a novel Energy-Predictive Drone Service (EPDS) framework for efficient package delivery within a skyway network. The EPDS framework incorporates a formal modeling of an EPDS and an adaptive bidirectional Long Short-Term Memory (Bi-LSTM) machine learning model. This model predicts the energy status and stochastic arrival times of other drones operating in the same skyway network. Leveraging these predictions, we develop a heuristic optimization approach for composite drone services. This approach identifies the most time-efficient and energy-efficient skyway path and recharging schedule for each drone in the network. We conduct extensive experiments using a real-world drone flight dataset to evaluate the performance of the proposed framework.

\end{abstract}


\begin{highlights}
  \item An Energy-Predictive Drone Service (EPDS) framework to minimize the average delivery time.
    \item An energy consumption prediction module for a longer time prediction using a shorter time sequence.
    \item A heuristic-based optimization for drone services composition to reduce recharging and waiting time. 
\end{highlights}


\end{frontmatter}


\section{Introduction}



The Internet of Things (IoT) has become more mature and widespread, largely thanks to advancements in software and hardware technologies. A drone (\textit{aka} Unmanned Aerial Vehicle (UAV)) is a type of IoT device that can fly autonomously (\cite{bine2024flavors}). Drones serve various purposes, including aiding in farm irrigation, capturing aerial imagery for entertainment, and facilitating the delivery of retail goods (\cite{mohsan2023unmanned}). Drone delivery services are increasingly important because they can offer faster delivery times, lower operational costs, and potentially a greener alternative to traditional delivery methods (\cite{eskandaripour2023last}).


Several key challenges, however, hinder the wider adoption of drones for delivery services (\cite{sah2021analysis}). A primary challenge is constrained battery capacity, which limits a drone's flight range (\cite{huang2022drone}). With current lightweight batteries, delivery drones are not well-suited for long-distance trips, particularly when carrying heavy payloads. As a result, some studies propose using drones only for last-mile deliveries (\cite{garg2023drones}). Despite these limitations, drones remain a clean, cost-effective, and ubiquitous alternative to land-based delivery in both urban and rural areas (\cite{attenni2023drone}).


Another critical challenge is \emph{predicting a drone’s energy consumption} in highly congested and dynamic environments. The main factors affecting drone energy consumption can be classified into four categories: drone design, operating environment, drone dynamics, and delivery operations (\cite{ZHANG2021102668}). These factors include the drone’s weight and battery capacity (design), weather conditions such as wind speed and air density (operating environment), flight speed and angle (dynamics), and payload weight, along with the number of delivery stops (operations). All these factors contribute to uncertainty in predicting drone energy consumption. Existing studies often model energy consumption as a linear function of time or distance (\cite{huang2020reliable}). While some approaches combine aerodynamic factors and drone design parameters (\cite{jayakumar2024design}) or use regression models based on specific flight scenarios (\cite{alyassi2022autonomous}), none of these approaches comprehensively account for all such factors \emph{or} provide near-future predictions of drone energy consumption. 


\emph{Service-Oriented Computing (SOC)} offers a promising approach to managing these complexities. SOC is a paradigm that treats services as fundamental constructs to enable rapid, cost-effective, and simple composition of distributed applications (\cite{yangui2021future}). Drones are a prime example of distributed applications perfectly suited to the service paradigm, exhibiting both functional and non-functional (\textit{aka} Quality of Service (QoS)) characteristics (\cite{alkouz2021service}). The primary functional characteristic of a drone delivery service is the successful transport of a package from a source to a destination. Non-functional characteristics include delivery time, cost, and energy consumption. To facilitate drone delivery services, we utilize a \emph{skyway network} infrastructure as the delivery medium, which enables drones to operate while avoiding restricted no-fly zones (\cite{bhandary2025predictive}).

A skyway network consists of interconnected nodes and the skyway paths between them, where each directly connected node pair represents a feasible flight segment. Each node serves as both a delivery location and a recharging station, equipped with mechanisms for rapid battery replenishment (e.g., Quick battery-Charging Machines (QCMs)) to support the drones’ limited battery capacities (\cite{qin2021multiobjective}). A significant challenge in such networks is avoiding congestion at QCMs; when multiple drones need to recharge concurrently, additional waiting time is incurred (\cite{al2021dncs}). Typically, a drone’s precise recharging duration becomes known only upon its arrival at a station. However, if the drone’s energy consumption for upcoming flight segments can be \textit{predicted} in advance, its recharging needs can be estimated ahead of time. This predictive approach enables proactive scheduling, which can significantly reduce waiting times at recharging stations.


We propose a machine learning-driven, \emph{energy-predictive service composition} approach for drone delivery. This approach prioritizes selecting flight paths that minimize congestion and waiting times. The approach comprises two main modules. The first module handles \textit{drone data collection} and \textit{communication} and focuses on gathering flight data (e.g., speed, battery voltage, payload) from drones. The second module, a \emph{smart control center}, receives and analyzes this data to predict energy consumption for each flight segment (i.e., a direct path between two nodes in the skyway network). Based on these predictions, the smart control center estimates the required recharging time for each drone at upcoming intermediate stations and can reschedule the takeoff or arrival times of other drones to minimize potential congestion. We assume the smart control center comprises edge nodes strategically distributed across the network to reduce communication latency. A heuristic algorithm is employed to guide drones along their routes, thereby reducing waiting times due to congestion and supporting continuous delivery, even when intermediate recharging is needed. We evaluate our approach using a real drone flight dataset collected from a 3D-modeled city CBD skyway network. Experiments are conducted on skyway networks of various sizes to assess the performance of the proposed algorithms in achieving optimal drone deliveries. More specifically, the main contributions of this paper are as follows:

\begin{itemize}
    \item We develop an EPDS composition framework that aims to minimize the average delivery time for drone services.
    \item We design an energy consumption prediction module that uses short historical time sequences to predict drone energy usage over longer future horizons accurately.
    \item We propose a heuristic optimization algorithm for drone service composition that proactively reduces both recharging delays and waiting times due to congestion. 
\end{itemize}

\subsection{Motivation Scenario}


\begin{figure}[t]

\includegraphics[width=0.79\linewidth]{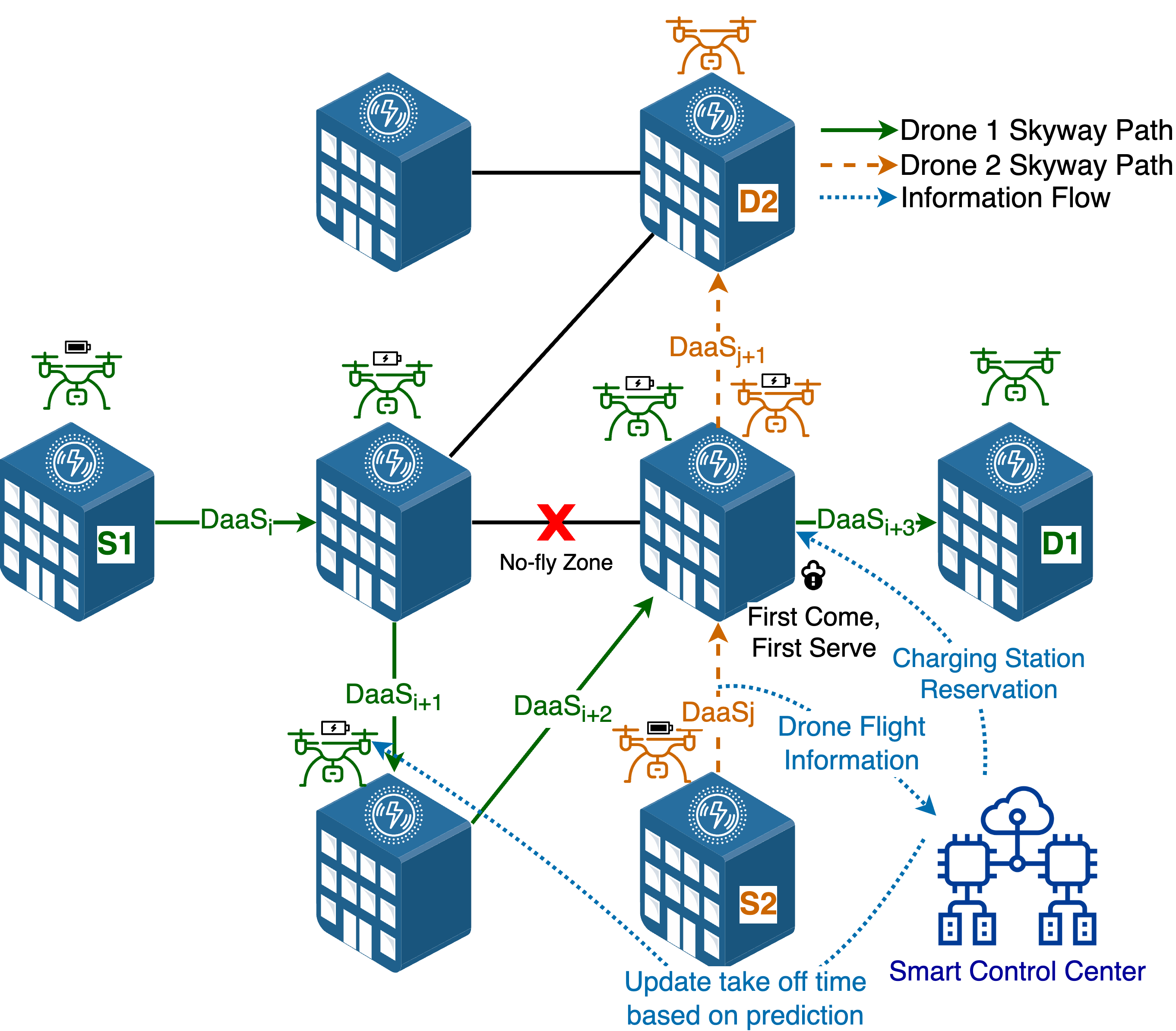}
\centering
\caption{Skyway Network with Multiple Drone Services}
\label{fig:scenario} 
\end{figure}

Consider an Illinois-based drone delivery service provider, as the state is well-suited for drone delivery activities (\cite{aurambout2022drone}). A customer requests a drone delivery from Chicago to Aurora (a distance of $\sim$83~km) to be delivered as quickly as possible. However, typical delivery drones have a maximum range of only 3-33~km ({\cite{10528255}}). This long distance, therefore, necessitates multiple intermediate recharging stops to complete the journey. Suppose a skyway network is constructed in accordance with Federal Aviation Administration (FAA) drone flight regulations\footnote{\url{https://www.faa.gov/uas/getting_started/where_can_i_fly}}. As shown in Figure~\ref{fig:scenario}, this network includes designated corridors while respecting restricted areas and no-fly zones. Each node in the network corresponds to the rooftop of an existing building in the Chicago area and is equipped with a limited number of drone recharging pads. Each such node functions as both a delivery point and a recharging station.

We envision a scenario where multiple drone service providers utilize this same network. This multi-tenant usage could lead to congestion at intermediate nodes and cause drones to queue for sequential recharging. Figure~\ref{fig:scenario} illustrates such a situation: Drones 1 and 2 both plan to use the same intermediate node, potentially causing a recharging bottleneck. One way to alleviate this is to adjust the takeoff times of some drones (i.e., reschedule their departure) to stagger their arrivals at the busy station. However, a naive (reactive) approach would wait until a drone actually arrives at a recharging station to gather information (e.g., its battery status and required recharging time) before making adjustments for others. Such reactive reallocation after drones have already reached a congested node is time- and energy-consuming.

We propose instead a \emph{proactive prediction} approach that estimates each drone’s energy consumption and needed recharging time early during the drone’s flight segment. By predicting a drone’s energy consumption well before it reaches the next station, we can estimate how long it will take to recharge at that station. Consequently, the system (smart control center) can proactively adjust the takeoff times of other drones still en route or waiting to depart. As shown in Figure~\ref{fig:scenario}, the smart control center updates the takeoff time of Drone~1 based on the in-flight status of Drone~2 and prevents them from overlapping at the recharging node. This proactive strategy can significantly reduce waiting times due to congestion and lead to faster overall delivery. Accurate energy consumption prediction is essential for this process to avoid recharging station congestion effectively.

We aim to design a \textit{heuristic-based energy-predictive} drone delivery framework driven by machine learning algorithms. Our approach predicts drone energy consumption by taking into account all key factors (such as wind conditions and payload weight) that influence a drone’s energy consumption. We model how each factor contributes to changes in the drone’s battery voltage; for example, carrying a heavier payload causes a faster voltage drop due to increased power demand. Using the predicted energy consumption, our framework computes optimal takeoff times for drones. If the system knows in advance how long a reserved recharging pad will be occupied, a drone can schedule its departure such that it arrives exactly when the pad becomes free. This ensures efficient utilization of recharging stations and avoids unnecessary queuing. Using this heuristic-based approach, we aim to optimize waiting times and compute an allocation plan for every drone operating in the network.

\section{Related Work}

Our proposed framework brings together concepts from two research areas: (1) drone delivery path planning and (2) drone energy consumption prediction and modeling. To the best of our knowledge, no existing work has yet introduced a similar \emph{energy-predictive} drone service framework.

\subsection{Drone Delivery}

A Drone Navigation and Charging Station (DNCS) algorithm is proposed to optimize drone routing by minimizing both total travel distance and battery charging costs while avoiding restricted no-fly zones (\cite{al2021dncs}). The objective of the study is to enhance drone trajectory planning by simultaneously accounting for multiple dynamic factors, including variable charging costs across different Quick Charging Machines (QCMs), queuing delays, and geographical constraints such as no-fly zones. The proposed shortest-path algorithms are based on modifying Dijkstra’s algorithm to incorporate charging costs and dynamically reroute drones around circular no-fly areas when necessary. Key components of the algorithm include a flight-time estimation model based on battery capacity, discharge rate, and payload-induced amperage draw; a cost-sensitive charging station selection mechanism; and a geometric path adjustment strategy for no-fly zones. A limitation of the study is that it assumes ideal communication conditions and homogeneous drone types, and it does not incorporate environmental factors such as wind effects.



A resilient drone delivery service framework is designed that explicitly considers wind conditions (\cite{SHAHZAAD2021335}). The study indicates that dynamic real-world environmental changes (like wind gusts) can significantly increase the failure rate of drone deliveries. A simple mathematical model is used to examine the wind speed and direction on drones. A resilient service composition with a self-adaptive lookahead mechanism is proposed that adapts to changing wind patterns during flight. However, this framework does not account for drone energy consumption when planning the delivery.

An energy-efficient drone delivery path planning framework is proposed using a reinforcement learning technique (\cite{hong2021energy}). The state-of-the-art Twin-Delayed Deep Deterministic Policy Gradient (TD3) algorithm is modified for the drone delivery context. The modified TD3 approach factors in key influences on energy consumption, including wind speed, wind direction, and payload weight, and learns to find the most energy-efficient path from source to destination \emph{without} using intermediate stations. The power consumption model in this framework is designed to fit instantaneous power usage (based on immediate feature inputs) rather than cumulative energy over a trip. The simulation experiments are performed on a small network and do not guarantee the scalability of the proposed approach.





\subsection{Drone Energy Consumption Modeling}

Modeling drone energy consumption is critical for planning optimal drone delivery services. Some prior works rely on using mathematical models to present drone energy consumption. For example, Huang et al.~(\cite{huang2020reliable}) present a drone delivery system that leverages a public transportation network to extend the delivery range. They use a stochastic model to estimate path travel times for a reliable routing solution and approximate drone energy consumption as a linear function of travel time. Additionally, they assume that energy usage during hover is roughly equal to that during forward flight. These simplifications make the model imprecise. Furthermore, there is no suggested energy consumption rate for this model.

Deng et al.~(\cite{deng2022vehicle}) propose a vehicle-assisted drone delivery scheme that allows drones to serve multiple customers in a single take-off while accounting for energy consumption based on varying payloads. The study aims to improve delivery efficiency by minimizing total service time, which includes both drone delivery time within customer clusters and vehicle travel time. To achieve this, three key models are introduced: a drone energy consumption model, a multi-drone task allocation model, and a vehicle path planning model. These models are solved using a hybrid heuristic algorithm that combines an improved K-means clustering method with ant colony optimization. The proposed scheme is based on several simplifying assumptions, including homogeneous drones and fixed service times. Although it takes into account payload weight, it does not consider other important factors, such as wind and weather conditions.

A genetic algorithm is proposed to optimize drone routes for energy efficiency (\cite{9098989}). The drone delivery routing is formulated as a Traveling Salesman Problem (TSP). The genetic algorithm’s strategy to reduce energy consumption includes minimizing the number of turns a drone makes. The energy model takes into account multiple factors, including velocity, acceleration, and deceleration, and whether the drone is flying straight or turning. Different equations are derived for different flight phases. While this model incorporates more parameters and aligns better with real-world energy consumption than simpler models, it still does not capture the full range of real-world uncertainties in drone operations.




A data-driven approach using Long Short-Term Memory (LSTM) neural networks is introduced to model drone energy consumption (\cite{muli2022comparative}). The LSTM-based model is compared against five benchmark energy models derived from physics-based considerations. Using sequences of real flight data as input features, the model predicts the drone’s energy consumption for a flight. The experimental results demonstrate that the machine learning approach outperforms traditional mathematical models in terms of prediction accuracy. It is found that the Bidirectional LSTM (Bi-LSTM) yields improved learning compared to LSTM by incorporating both past and future contexts during training. Energy usage depends on past states (e.g., speed, wind resistance, payload) and also on upcoming actions such as turns, climbs, or deceleration near recharging stations. A limitation of this study is that while it models energy consumption from input sequences, it does not explicitly use the model to predict a drone’s future power consumption over a route.

\section{Energy-Predictive System for Drone Services}

We propose an energy-predictive system for drone services. Our system scenario involves multiple drones providing delivery services to their respective destinations within a skyway network of landing and recharging stations (see Figure~\ref{fig:scenario}). When a drone is fully recharged, its battery is enough to cover each segment flight in the skyway network. When a customer invokes a drone delivery service, a corresponding service request is created. The smart control center collects and analyzes flight data from all drones, predicts energy consumption for each segment based on wind, payload, and other relevant factors, allocates the availability of recharging stations, and assigns paths and priorities to each drone service. We define that each drone needs to be fully recharged at the recharging stations. Based on predicted energy consumption and flight times, the smart control center computes the right takeoff times for each drone service in advance.

We formally define the key concepts of our framework, including energy-predictive drone service, service request, composite service, and recharging station congestion.

\begin{definition}\textbf{Energy-Predictive Drone Service EPDS.}
An EPDS represents the delivery functionality of a drone carrying a package from a source node to a destination node in the skyway network, augmented with timing attributes and QoS constraints derived from energy consumption predictions. Formally, an EPDS is a tuple $\langle \mathit{EPDS}_{id}, EPDS_f, EPDS_q \rangle$, where:
\begin{itemize}
    \item $\mathit{EPDS}_{id}$ is a unique identifier for the drone service.
    \item $EPDS_f$ is the drone delivery function based on Energy Consumption Prediction (ECP), consisting of:
    \begin{itemize}
        \item $\langle Loc_{src}, Loc_{des} \rangle$: the pickup (source) and delivery (destination) locations (nodes in the skyway network).
        \item $\langle V_0, V_1, \dots, V_n \rangle$: the sequence of the drone’s battery voltages recorded during flight, sampled every 100~ms.
        \item $\langle T_{src}, T_{flight}, T_{des} \rangle$: a tuple of time attributes, where $T_{src}$ is the drone’s takeoff time at the source, $T_{flight}$ is the estimated flight duration for this EPDS, and $T_{des}$ is the landing time at the destination. The total delivery time $T_t$ for this drone service is $T_{des} - T_{src}$.
    \end{itemize}
    \item $EPDS_q$ is a tuple of QoS parameters $\langle q_1, q_2, \dots, q_m \rangle$ for the service, where each $q_i$ represents a non-functional attribute (e.g., flying speed, battery capacity, etc.).
\end{itemize}
\end{definition}

\begin{definition} \textbf{EPDS Request ER.}
    An ER denotes a customer’s delivery request. It is defined as a tuple $\langle S, D, P \rangle$, where $S$ is the pickup location, $D$ is the delivery location, and $P$ is the package weight.
\end{definition}


\begin{definition} \textbf{EPDS Composite Service ECS.}
Given a set of EPDSs $S_{EPDS} = \{\mathit{EPDS}_1, \mathit{EPDS}_2, \dots, \mathit{EPDS}_n\}$, we define an ECS as an orchestration of multiple EPDSs to fulfill a delivery request that spans multiple segments. An ECS is a tuple $\langle \mathit{ECS}_{id}, S_{EPDS}, Path, ER \rangle$, where:
\begin{itemize}
    \item $\mathit{ECS}_{id}$ is a unique identifier for the composite service.
    \item $S_{EPDS}$ is the set of constituent EPDSs $\{\mathit{EPDS}_1, \mathit{EPDS}_1,\dots,\mathit{EPDS}_n\}$.
    \item $Path$ is the overall skyway route from the source to destination, formed by chaining the segments of the EPDSs in $S_{EPDS}$. If the skyway network nodes along this route are $N_1 \rightarrow N_2 \rightarrow \cdots \rightarrow N_{n+1}$, then each $\mathit{EPDS}_i$ corresponds to the segment between $N_i$ and $N_{i+1}$.
    \item $ER$ is the EPDS Request that this composite service is satisfying.
\end{itemize}
\end{definition}

\begin{definition} \textbf{Recharging Station Congestion RSC.}
An RSC event represents the scenario where multiple drones simultaneously request recharge at the same station. Formally, we define RSC as a tuple $\langle S_{ER}, S_{ECP}, T_c \rangle$, where:
\begin{itemize}
    \item $S_{ER}$ is the set of EPDS requests (deliveries) that have at least one recharging station in common along their planned skyway paths.
    \item $S_{ECP}$ is the set of energy consumption prediction outcomes (one for each EPDS in $S_{ER}$) relevant to the shared station.
    \item $T_c$ is the full recharge time at that station (i.e., the time required for a drone's battery to recharge from 0\% to 100\%).
\end{itemize}
\end{definition}

Table~\ref{table:notation} summarizes the notations used in our formal definitions.

\begin{table}[t]
\centering
\caption{Summary of Notations}
\begin{tabular}{ll}
\hline
\textbf{Symbol} & \textbf{Meaning} \\
\hline 
$EPDS$ & Energy-Predictive Drone Service \\
$Loc_{i}$ & A node (location) in the skyway network \\
$V$ & Battery voltage of a drone \\
$T_t$ & Total time of a drone delivery service \\
$ER$ & EPDS Request (delivery request) \\
$RSC$ & Recharging Station Congestion \\
$T_c$ & Full recharge time at a station (from 0\% to 100\%) \\
$ECP$ & Energy Consumption Prediction \\
\hline
\end{tabular}
\label{table:notation}
\end{table}

\subsection*{Problem Formulation}

Given a set of energy-predictive drone service requests from customers, our problem can be formulated as finding the optimal composition of $EPDSs$ to deliver packages from their pickup locations $S$ to their respective destinations $D$. The goal is to leverage $ECP$ to avoid recharging station congestion $RSC$ and minimize the overall delivery time.

The composition of drone delivery services must satisfy several constraints: First, intrinsic drone constraints such as each drone’s weight, flight speed, payload capacity, and battery capacity must be respected when composing services. Second, extrinsic environmental constraints, such as wind speed and wind direction, can affect flight segments and must be considered when composing services. Third, the availability of recharging stations is limited: if multiple drones plan to use the same station at overlapping times, an RSC situation occurs, potentially increasing the average delivery time significantly.

As discussed earlier, most existing studies either apply mathematical modeling or machine learning to drone energy consumption, but to our knowledge, none predict drones’ energy consumption in a skyway network setting to aid service composition. There are several key challenges in composing $EPDSs$ optimally: (1) Avoiding $RSC$ when multiple drones target the same recharging station around the same time; (2) Selecting the right machine learning model and effectively converting raw flight data into accurate energy-consumption predictions to be used for optimization; (3) Handling the varying flight distances of different $EPDS$ segments under diverse conditions, which makes it non-trivial to choose a fixed prediction horizon or uniform input sequence length for energy prediction; (4) Conducting real-world experiments to validate energy consumption under different conditions, which is essential for ensuring the prediction model’s accuracy.

Given the above constraints and challenges, our objective is to find the optimal composition of $EPDSs$ by addressing two aspects: (i) accurately predicting the energy consumption of drone services under varying extrinsic conditions, and (ii) composing $EPDSs$ using a machine learning-driven heuristic to minimize the total delivery time $T_t$. We focus on optimizing the initial service composition plan and then dynamically refining it using our predictions, in order to complete all deliveries in the minimum time.


\section{Energy-Predictive Drone Service Framework}

\begin{figure}[t]

\includegraphics[width=\linewidth]{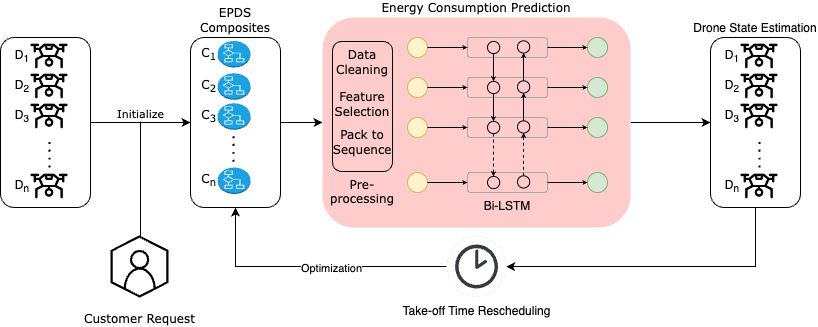}
\centering
\caption{Overview of EPDS Composite Framework}
\label{fig:framework} 

\end{figure}

Our proposed Energy-Predictive Drone Service (EPDS) framework comprises two main modules. Figure \ref{fig:framework} provides an overview of these two modules. The first module is the energy prediction module, which trains a prediction model at each node using drone flight data. The second module takes the energy consumption prediction, i.e., the output of the first module, and schedules the takeoff of concurrently operating services. This module aims to reduce the average delivery times of all concurrent operating services. We assume a central smart control center is responsible for doing this planning to reduce the overall congestion in the skyway network.



\subsection{Initial Composition}

An EPDS composition aims to compose the best path, i.e., the shortest delivery time, for a customer request from the source node to the destination node. Each node in the skyway network is assumed to have limited availability of recharging pads. We formally define a skyway network Node as follows:


\begin{definition} \textbf{Node N.}
    A Node is a building rooftop in a skyway network with takeoff, landing, and recharging functions for drones. N is a tuple of \textlangle $N_{id}, N_{status}, Neighbors$\textrangle, where
    \begin{itemize}
        \item $N_{id}$ is a unique identifier for a node.
        \item $N_{status}$ represents the availability status of the recharging pad. $N_{status}$ is a tuple of \textlangle $T_{start}, T_{end}, Status$\textrangle, where $T_{start}$ and $T_{end}$ are the starting time and ending time of the status and $Status$ can only be \textit{recharging} or \textit{pred\_recharging}. The \textit{recharging} means that the node is being used for recharging, and \textit{pred\_recharging} means that the node is reserved for a period of time based on the energy consumption prediction.
        \item $Neighbors$ is a set of neighbor nodes that are connected to this node.
    \end{itemize}

\end{definition}

A skyway network is a fully connected spatiotemporal network with many Nodes and edges. Given a service request $ER$, we first find the best path based on various composition strategies. We use a distance-based A-star heuristic approach as our baseline strategy. The heuristic function is the distance between two nodes. Another comparison strategy is the Brute-Force approach, which searches all possible composition plans and finds the optimal composition. Our composition strategy is an \textit{energy} and \textit{distance} aware heuristic approach, which differs from A-star through our own defined heuristic function. Moreover, the time it takes for a drone to recharge should be taken into account, and the total delivery time can be calculated as the sum of waiting time, flight time, and recharging time. In order to equalize the influence of distance and energy, we normalize them by the drone's speed and recharge rate. The heuristic function is defined as follows:


\begin{equation}
h(EPDS)=\frac{D(Loc_{i}, Loc_{des})}{V}+\frac{E(Loc_{i}, Loc_{des})}{Rate_{recharge}}\label{hfunc}
\end{equation}

where $D$ is a Euclidean−distance function calculating the line-of-sight from the current node to the final destination, $E$ is the energy estimation function from the current node to the destination without wind impact and payload, $Loc_{i}$ and $Loc_{des}$ are the current and destination nodes respectively, $V$ is the drone speed, and $Rate_{recharge}$ is the recharge rate.

For each $ER$, we initialize an optimal composition plan process. The initial composition plan includes not only path planning but also time planning. The algorithm schedules each drone according to the First-Come, First-Served (FCFS) strategy. In this study, we do not incorporate priority rules or exceptions to the FCFS policy, as our focus is on baseline scheduling performance. Existing works have explored such priority rules or preemption (\cite{alkouz2021service,sung2020zoning}). In this respect, FCFS means that the drone that arrives at the recharging station first has a higher priority for taking off. For example, given an $ER$ \textlangle $S_{1}, D_{1}, P_{1}$ \textrangle, we first find the optimal path between $S$ and $D$ through the proposed heuristic approach. Because there is only one $ER$, the time can be planned simply with this drone as the top priority. The initial composition plan will then be assigned to the drone to deliver the package. However, if there is another $ER$ \textlangle $S_{2}, D_{2}, P_{2}$ \textrangle, a composite service plan is needed to solve multiple $ER$s. Using the proposed heuristic approach, the algorithm first finds the optimal path for $ER_{1}$ and $ER_{2}$. For example, let's say we find $Path_{1} = \textlangle S_{1}, A, B, D_{1} \textrangle$ and $Path_{2} = \textlangle S_{2}, A, C, D_{2} \textrangle$ to be the optimal paths for two $ER$s.  As can be seen, a recharging station congestion $RSC$ is detected at node $A$, implying that congestion could occur at $A$ because it is a shared node in both paths. Assuming node $A$ is closer to node $S_{1}$ than node $S_{2}$, the first-come, first-served policy, will give $Path_{1}$ a higher priority. A new takeoff time $T_{src}$ is assigned to the first EPDS composition plan $C_{1}$ while the time attributes for the other composition plans remain pending and will be updated during the delivery. As a result, the EPDS initial composite service plans \textlangle $C_{1}, C_{2}$\textrangle composed. These composites take no account of real-time energy consumption predictions. They are solely based on the original path and estimated energy consumption.


\subsection{Energy Consumption Prediction}
Given a set of pre-planned drone services for different service providers, this module predicts the actual energy consumption of each service at the next node. Since the prediction uses real in-flight data, it considers intrinsic and extrinsic constraints like wind conditions. The output of this module will be used to update the takeoff time of drones that use the same nodes to reduce the waiting time. For simplicity, we assume the original composite services from the source to the destination for each drone provider are not changed, but rather the takeoff time from the node is altered, i.e., $T_{src}$ and $T_{des}$.

The prediction module mainly depends on the collected in-flight data. Therefore, we design a real-world experiment (see Section \ref{data}) to collect the energy consumption history data of multiple drones under different wind conditions. Table~\ref{table:attributes} illustrates this dataset's attributes (features). 
\begin{table}[t]
\caption{Attributes in the Dataset}
\begin{tabular}{c|c}
\hline
\textbf{Attribute Name}&{\textbf{Description}} \\
\hline
$T$ &  Time passed in milliseconds\\
$ES_{x/y/z}$ & Current estimation for x, y, z coordinates\\
$ST_{roll/pitch/yaw}$ & Current estimation for roll, pitch, yaw\\
$Vbat$ & Current estimation for voltage\\
$Wind_{speed}$ & Speed of wind: km/hours\\
$Wind_{direction}$ & Global and relative wind directions\\
$Dis$ & Distance the drone has traveled so far\\
$Loc_{role}$ & Role of the current node, i.e., start, fly, destination.\\
$D_{id}$ & ID of the drone\\
$Loc$ &  Name of the nodes in the skyway path network\\
$Len_{pred/in}$ & Prediction and input windows sizes\\
\hline
\end{tabular}
\label{table:attributes}
\end{table}

Given all the features in the dataset, we formally define a feature set as follows:

\begin{definition} \textbf{Feature Set}
    A feature set is a tuple of \textlangle $ES_{x/y/z}$, $ST_{roll/pitch/yaw}$, $Vbat$, $Wind_{speed}$, $Wind_{direction}$, $Dis$, $Loc_{role}$\textrangle
\end{definition}

The prediction module is further divided into two main steps. The first is data preprocessing, which prepares the data for prediction. The second is the Bi-LSTM model that predicts energy consumption. In what follows, we describe each step.

\subsubsection{Data Preprocessing}
The data preprocessing step has three main tasks: data cleaning, feature selection, and sequence packaging. Given the data from drone flight information, the data is first cleaned and normalized into the required format. We employ various strategies to select the optimal set of features that yields accurate predictions. These strategies include selecting all features in the feature set, selecting only $Vbat$, or applying Principal Component Analysis (PCA) for feature selection. Based on the chosen feature selection strategy and the sequence window length, we then pack the data into sequences. A Feature Sequence (FS) is a sequence of the feature attributes selected from the feature set and can be an input or output sequence with a given sequence length ($Len_{in}$ or $Len_{pred}$).




\subsubsection{Bi-LSTM Prediction}

\begin{figure*}[t]
\centering
  \includegraphics[width=\linewidth]{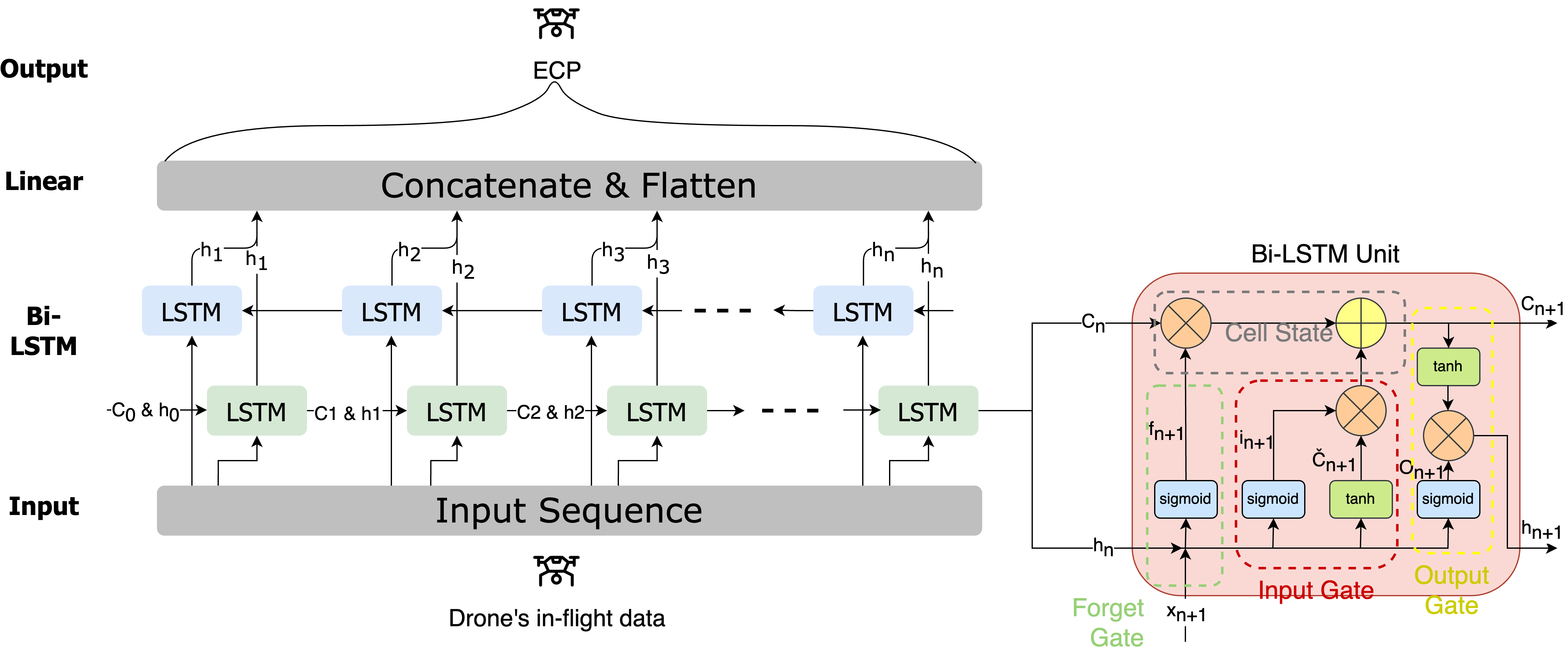}
  \caption{Example of Bi-LSTM}
\label{fig:Bi-LSTM}
\end{figure*}

We selected the Bi-LSTM model for the $ECP$ because it fits the nature of our dataset, which consists of high-frequency sequential flight data with complex temporal patterns influenced by wind, battery voltage, and flight dynamics. Its ability to learn bidirectional dependencies makes it well-suited for capturing both past and future context, which is essential for accurate energy prediction in drone operations. Figure \ref{fig:Bi-LSTM} presents an example of the Bi-LSTM architecture in our context. The input layer presents the drone's set of feature sequences. The output layer is the $ECP$, the total energy consumed at the next node. An LSTM unit (\cite{van2020review}) is equipped with three gates: an input gate, a forget gate, and an output gate to store the information. For an LSTM unit, the input is always the current input $X_{t}$, previous hidden state $h_{t-1}$, and previous cell output state $c_{t-1}$. Calculating the three gates in an LSTM unit follows the equations \eqref{ft}-\eqref{ot}, where $f$, $i$, and $o$ represent forget gate, input gate, and output gate, respectively.





\begin{equation}
f_{t} = \sigma(W_{f}[h_{t-1},x_{t}] + b_{f})\label{ft}
\end{equation}

\begin{equation}
i_{t} = \sigma(W_{i}[h_{t-1},x_{t}] + b_{i})\label{it}
\end{equation}

\begin{equation}
o_{t} = \sigma(W_{o}[h_{t-1},x_{t}] + b_{o})\label{ot}
\end{equation}

\begin{figure}[t]
\centering
\includegraphics[width=0.75\linewidth]{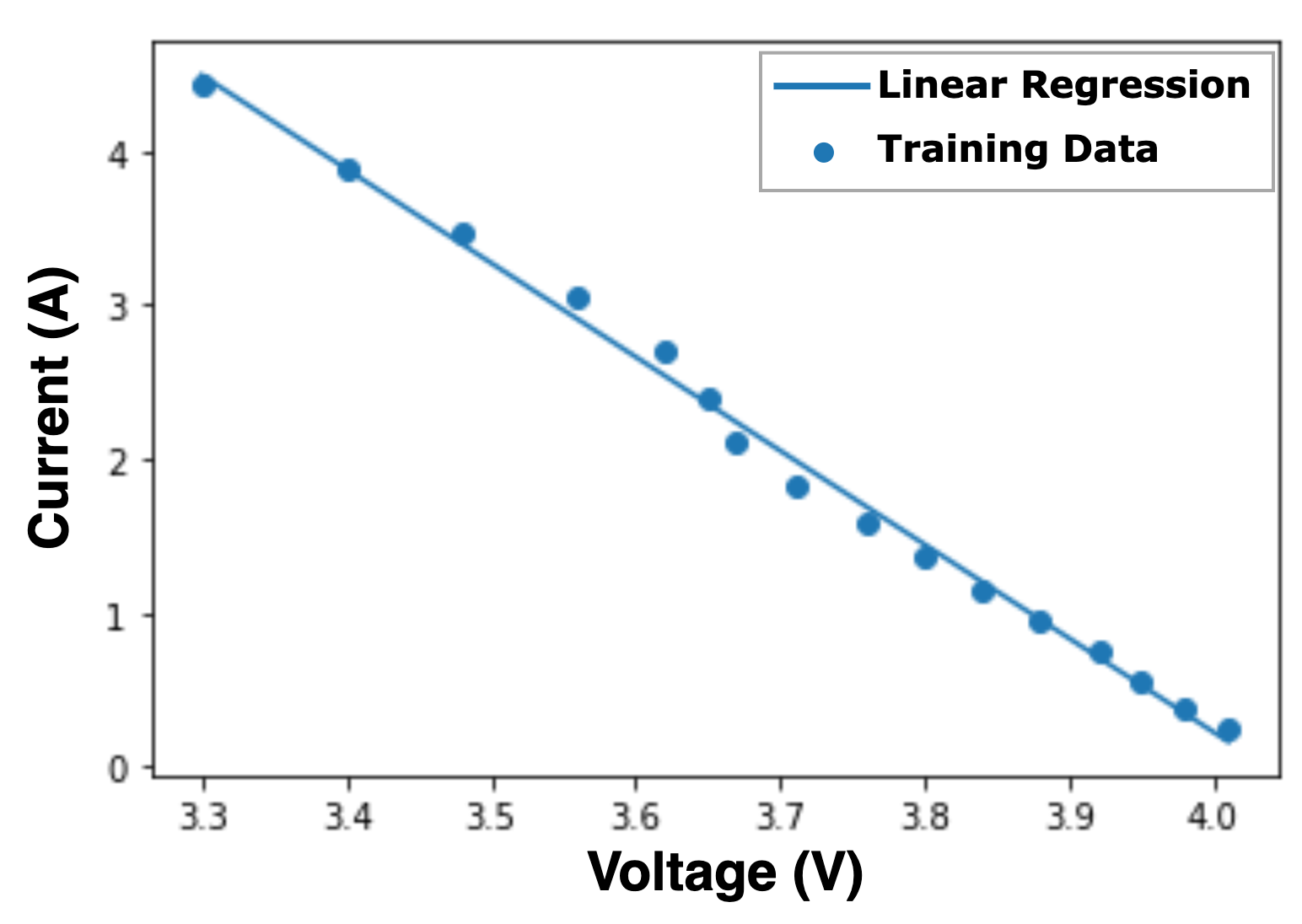}
\caption{Voltage to Current Conversion}
\label{fig:vtoa}
\end{figure}

In the equations above, $W$ represents the weights, $b$ is the bias, $*$ is the element-wise product, and $\tanh$ is the activation function. Given an FS at the input layer, the input sequence is in the shape of \textlangle $B$, $F$, $Len_{in}$ \textrangle, where $B$ is the batch size of the training or testing dataset, $F$ is a subset of features selected from the feature set, and $Len_{in}$ is the length of input window size. The hidden state $h_{t}$ is calculated using equations \eqref{cc}-\eqref{h}.

\begin{equation}
\tilde{C} = \tanh{(W_{c}[h_{t-1},x_{t}] + b_{c})}\label{cc}
\end{equation}

\begin{equation}
C_{t} = f_{t}*C_{t-1} + i_{t}*\tilde{C}\label{C}
\end{equation}

\begin{equation}
h_{t} = o_{t}*\tanh{C_{t}}\label{h}
\end{equation}

where $\tilde{C}$ is the candidate for updating cell state, $C_{t}$ is the new cell state. In a Bi-LSTM model, one more step is involved in updating the hidden state through Equation \eqref{bt}.

\begin{equation}
h_{t} = [\overrightarrow{h_{t}}^\frown\overleftarrow{h_{t}}]\label{bt}
\end{equation}

where $^\frown$ represents the concatenation of two sequences, $\overrightarrow{h_{t}}$ and $\overleftarrow{h_{t}}$ are the hidden states of two directions (forward and backward). Last, the hidden states are flattened through a linear layer, and the prediction sequence is in the shape of \textlangle $B$, $Len_{pred}$ \textrangle. The prediction sequence does not contain the feature set's features but only $Vbat$. However, $Vbat$ cannot represent the energy consumption directly. Thus, we convert the prediction sequence to the final energy consumption $Q_{Energy}$. The Sum of $Vbat$ can represent the $Q_{Energy}$ using equations \eqref{eq1} - \eqref{eq2}.




\begin{equation}
I_{current}=func(Vbat)\label{eq1}
\end{equation}

\begin{equation}
Q_{Energy}=\sum(I_{current}*d_{Time})\label{eq2}
\end{equation}

where $I_{current}$ is the current, and $func()$ is the linear regression function converting voltage to current \footnote{\url{https://www.bitcraze.io/documentation/repository/crazyflie-firmware/master/functional-areas/pwm-to-thrust/}} as shown in Figure~\ref{fig:vtoa}. $Vbat$ is the voltage, $Q_{Energy}$ is the energy consumption, and $d_{Time}$ is the time interval, which is 100 ms in our experiments.

The lengths of the skyway segments are variable in real-world settings. Therefore, we cannot use a fixed prediction length for the energy consumption prediction. To address this issue, we devise an algorithm that predicts energy consumption for various segment lengths. Algorithm \ref{alg:length} describes the steps involved in predicting energy consumption with variable prediction lengths. First, the algorithm selects the feature attributes from the feature set to acquire the input sequence $Seq_{in}$ from the drone flight history at an early stage. $Seq_{in}$ is the input to the algorithm. Then, the total sequence length $L$ is initialized to 0 (Line \ref{l:line:initialize}). While the total sequence length $L$ is less than the length of the segment flight, the energy consumption prediction process is repeated using Bi-LSTM (Lines \ref{l:line:while begin} - \ref{l:line:while end}). The algorithm predicts the energy consumption over the previous prediction (Line~\ref{l:line:bi}) and concatenates all the output sequences as the final output (Line~\ref{l:line:concat}). The input sequence $Seq_in$ and total output sequence length $L$ are also updated at every step (Lines \ref{l:line:seq} - \ref{l:line:L}). If the output sequence length $L$ is larger than the segment flight $Len_{seg}$, the algorithm clips the output sequence from the beginning up to the length of the segment flight (Lines \ref{l:line:if begin} - \ref{l:line:if end}).

\begin{algorithm} [t]
 \caption{Energy Consumption Prediction with Variable Prediction Length} \label{alg:length}
 \small 
 \begin{algorithmic}[1] 
 \renewcommand{\algorithmicrequire}{\textbf{Input:}} 
 \renewcommand{\algorithmicensure}{\textbf{Output:}} 
 \REQUIRE $Seq_{in}$ 
 \ENSURE  $Out_{pred}$ 

 \STATE $L \gets 0$ \label{l:line:initialize}
 \WHILE{$L < Len_{seg}$} \label{l:line:while begin}
 \STATE $Seq_{pred} \gets Bi-LSTM(Seq_{in})$\label{l:line:bi}
 \STATE $Out_{pred} \gets {Out_{pred}}^\frown{Seq_{pred}}$\label{l:line:concat}
 \STATE $Seq_{in} \gets Seq_{pred}$\label{l:line:seq}
 \STATE $L \gets L + len(Seq_{pred})$\label{l:line:L}
 \ENDWHILE \label{l:line:while end}
 \IF{$L > Len_{seg}$}\label{l:line:if begin}
    \STATE $Out_{pred} \gets Out_{pred}[:Len_{seg}]$\label{l:line:if}
 \ENDIF\label{l:line:if end}
\end{algorithmic} 
\end{algorithm} 



\subsection{EPDS Composite Service Optimization} \label{timecomplxitysection}
The next stage is optimizing the composite services to reduce the overall time for completing all the delivery requests. As Figure \ref{fig:framework} shows, once the initial composition starts, our ECP model starts to work to predict the energy consumption for the current EPDS. Then,  the drone state is estimated based on the prediction. For example, given an EPDS composite service \textlangle $ECS_{1}, ECS_{2}$\textrangle, we first pop out $ECS_{1}$ to execute following the priority. The path for $ECS_{1}$ is $\textlangle S_{1}, A, B, D_{1} \textrangle$. When the drone $D_{1}$ takes off from the source node $S_{1}$, the smart control center starts collecting the drone flight information to analyze the energy consumption prediction module data. When $D_{1}$ has flown for a while, such as 20 percent of the segment flight from $S_{1}$ to $A$, the smart control center acquires an accurate energy consumption prediction $ECP$ for this segment flight. Based on the prediction, the recharging time for $D_{1}$ at node $A$ is estimated, and the availability time of recharging station $A$ gets updated. With the drone state estimation, the smart control center reschedules the takeoff time for other drones on the waiting list pending flight. As a result, $D_{2}$ gets informed about the time availability of node $A$ ahead of time. This leads to an earlier takeoff time since $D_{2}$ does not need to wait until $D_{1}$ arrives at node $A$. 


Algorithm \ref{alg:EC} presents the process of optimizing the EPDS Composite. In the algorithm, $H$ is our heuristic approach for path planning, $FCFS$ is a sorting function based on our First-Come First-Served policy, $T.Available$ is the available time of the recharging station, and $ECPModule$ is the energy consumption module that helps predict the energy consumption for a segment flight.  Given a set of ECSs $S_{ECS}$, a for loop initializes a path for every ECS in $S_{ECS}$ (Lines \ref{ec:lines:for begin} - \ref{ec:lines:for end}). Using the proposed heuristic approach, a path for each ECS is initialized using the acquired source and destination nodes from the ER (Lines \ref{ec:lines:get ER} - \ref{ec:lines:H}). The set of composites is then sorted based on the first-come, first-served policy (Line \ref{ec:lines:fcfs}). Then, for every ECS in the set, the takeoff time is optimized (Lines \ref{ec:lines:for begin1} - \ref{ec:lines:for end1}). For every ECS, the algorithm first acquires the next node (Line \ref{ec:lines:next node}) and sets the first available time of the next node minus the current drone's flight time ($ECS.EPDS.T_{flight}$) as the takeoff time (Line~\ref{ec:lines:takeoff}). We assume that the drone's flight speed is fixed so that the flight time can be computed. After $T_{takeoff}$, the algorithm starts collecting the drone's in-flight information, including voltage, speed, and flight dynamics, and then processes it into an input sequence $Seq_{input}$ (Line \ref{ec:lines:fly}). This input sequence explicitly ensures that the energy prediction module takes into account real-time battery consumption. Using the ECP module, the energy consumption is predicted for the current segment flight (Line \ref{ec:lines:ecp}). Based on the new drone state estimation (Line \ref{ec:lines:update drone}), the available time gets updated for the next node ahead of time (Line \ref{ec:lines:update time}). The newly available time will guide the other drones' takeoff times.

\begin{algorithm} [t]
 \caption{EPDS Composite Optimization ($S_{ECS}$)}\label{alg:EC}
 \small 
 \begin{algorithmic}[1] 
 \FOR{\texttt{ECS in $S_{ECS}$}} \label{ec:lines:for begin}
    \STATE $Src, Des \gets ECS.ER$\label{ec:lines:get ER}
    \STATE $ECS.path \gets H(Src, Des)$\label{ec:lines:H}
 \ENDFOR\label{ec:lines:for end}
 \STATE $S_{ECS} \gets FCFS(S_{ECS})$\label{ec:lines:fcfs}
 \FOR{\texttt{ECS in $S_{ECS}$}}\label{ec:lines:for begin1}
    \FOR{\texttt{N in $ECS.path$}}
        \STATE $N_{next} \gets next(ECS.path, N)$\label{ec:lines:next node}
        \STATE $T_{takeoff} \gets N_{next}.T_{Available}-ECS.EPDS.T_{Flight}$\label{ec:lines:takeoff}
        \STATE Collect flight data $Seq_{input}$\label{ec:lines:fly}
        \STATE $ECP \gets ECPModule(Seq_{input})$\label{ec:lines:ecp}
        \STATE Update Drone State Estimation $DSE$ based on $ECP$\label{ec:lines:update drone}
        \STATE $N_{next}.T_{Available} \gets DSE$\label{ec:lines:update time}
    \ENDFOR
 \ENDFOR\label{ec:lines:for end1}
\end{algorithmic} 
\end{algorithm}

In what follows, we discuss the time complexity of the three main modules proposed: the initial composition, the energy consumption prediction, and the EPDS optimization. We evaluate the effectiveness of our initial composition algorithm by comparing the time complexity to other algorithms, including the exhaustive search algorithm, Bellman-Ford (\cite{8847727}), and Dijkstra (\cite{luo2020surface}). 
For an exhaustive search algorithm, the time complexity is $O(N!)$ since it needs to search through all the possible paths. For Dijkstra and Bellman-Ford, the time complexity are $O(|E|+|V|Log|V|)$ and $O(|V|*|E|)$ respectively (\cite{abusalim2020comparative}). Our proposed initial composition algorithm is a heuristic approach modified from the A* algorithm; thus, the time complexity is $O(|E|)=O(b^{d})$ where $b$ is the branching factor of the graph, and $d$ is the depth of the destination node. For the second module, i.e., Bi-LSTM, the time complexity for training is $O(W)$ for every LSTM update, where $W$ is the number of weights (\cite{van2020review}). Last, for our EPDS optimization, the algorithm works with an $O(N*h*Max(h,d))$ complexity calculated from equations \eqref{ft} - \eqref{bt}, where $h$ is the hidden size and $d$ is the input size. Since the optimization only works during the flight while other drones are waiting, the time complexity of the EPDS optimization has little impact on the average delivery time.

\section{Multi-Drone Dataset}\label{data}

\begin{figure}[t]
\centering
\includegraphics[width=0.75\linewidth]{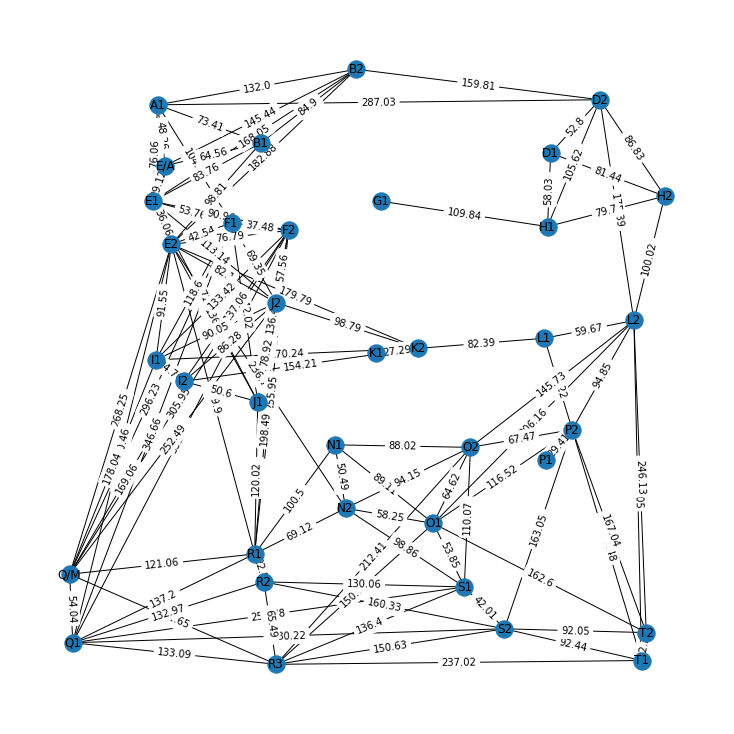}
\caption{Skyway Network}
\label{fig:skyway} 
\end{figure}

\begin{figure}[t]
\centering
\includegraphics[width=0.75\linewidth]{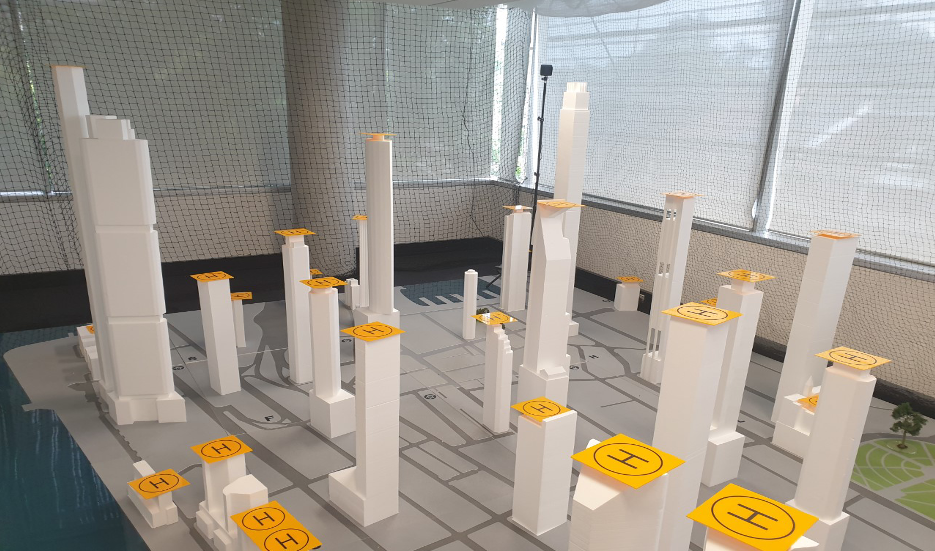}
\caption{Skyway Network 3D Model}
\label{fig:3D} 
\end{figure}

Using a real drone dataset is crucial to assess the performance of our proposed approach. We collected a real dataset of multiple drones operating in the same skyway network, as illustrated in Figure~\ref{fig:skyway}. This multi-drone dataset recorded various drone delivery operations, including waiting, flight, recharging, and arrival times under different wind speeds and directions. We conducted several experiments using three Crazyflie 2.1 nano-quadcopter drones to collect data on drone flight information, including position and voltage, as well as environmental information such as wind speed. We implemented a 3D model of the city's CBD area as our multi-drone skyway network, where the rooftops of high-rise buildings served as nodes for delivery and recharging (see Figure~\ref{fig:3D}). Each node was treated as both a delivery location and a recharging station, equipped with wireless charging pads on its rooftop. The system considered recharging availability constraints for long-distance drone delivery, where drones stopped at intermediate nodes to recharge while other drones remained waiting.

\begin{figure}[t]

\includegraphics[width=0.79\linewidth]{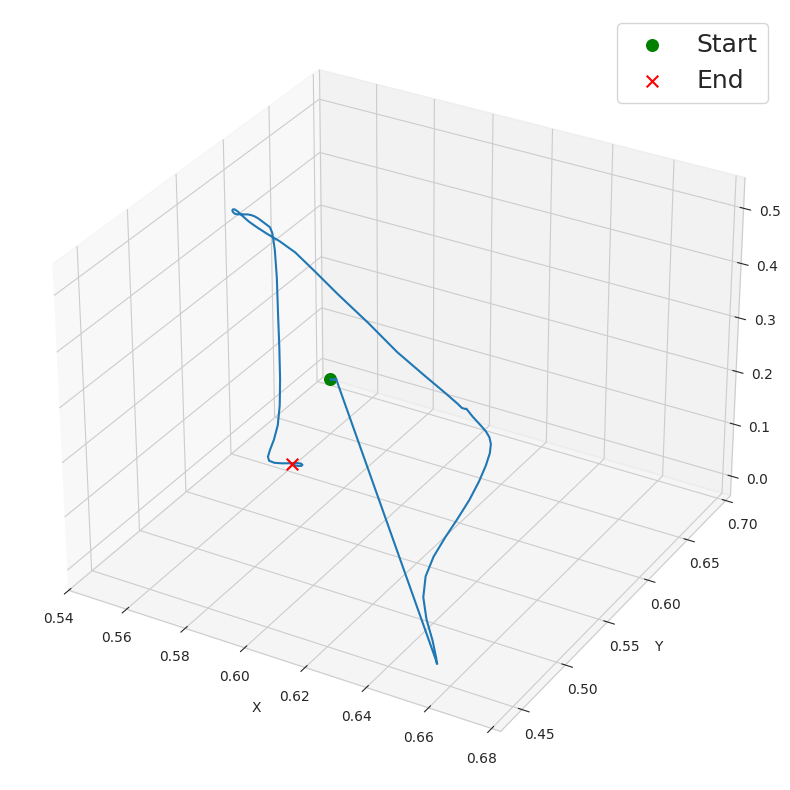}
\centering
\caption{Representative 3D Flight Trajectory}
\label{fig:traj} 
\end{figure}

\begin{figure}[t]

\includegraphics[width=0.79\linewidth]{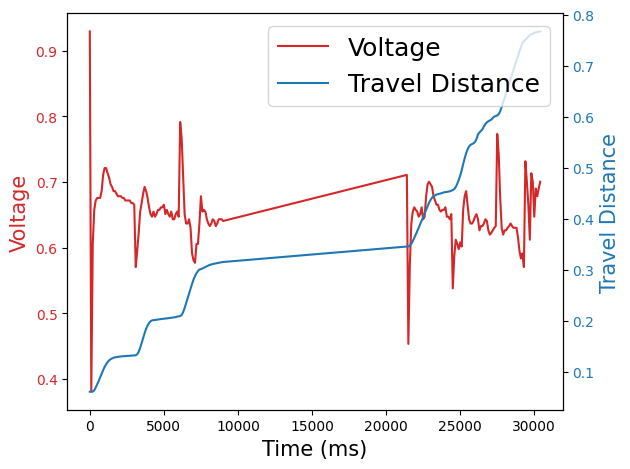}
\centering
\caption{Typical Current Draw Profiles}
\label{fig:current} 
\end{figure}

We used an adjustable fan to mimic the effects of different wind speeds under controlled indoor conditions. To calculate the coordinates of a drone, we used two HTC Vive base stations equipped with infrared laser tracking. The Crazyflie Python API was employed for data collection, with one or two intermediate recharging stations used as needed. We performed 70 flights to collect the following data attributes with a timestamp of 100 ms: (x, y, z) coordinates, flight dynamics (roll, pitch, and yaw), voltage history, wind speed and direction, travel distance, and waiting time at each recharging station. For each flight, the drone ascended vertically from the source node to a designated cruising altitude based on the elevation of the rooftop of the next destination node, then traveled horizontally at that altitude. Vertical ascents and descents were performed only at nodes rather than gradually during horizontal segments. Figure \ref{fig:traj} presents the 3D trajectory of one representative flight. We tested three levels of wind conditions: no wind, 6.1 km/h wind, and 7.6 km/h wind, representing typical low to moderate indoor wind speeds for lightweight drones. Each test flight was conducted under three different wind directions (North, South, and East) relative to the city’s 3D model, determined by the fan’s location. The wind vector magnitude was controlled by adjusting the fan speed setting, and the vector direction was set by positioning the fan relative to the drone’s planned path (\cite{bradley2023service}). When a drone arrived at a recharging station, the voltage increased to a threshold of 4.15 volts when fully recharged, which triggered the drone with the highest priority on the waiting list to take off. Figure \ref{fig:current} shows a representative current draw profile captured from our dataset for a complete drone flight and highlights how current consumption varies across phases such as takeoff, cruise, hover, and recharging. This empirical profile demonstrates that our energy prediction module is informed by realistic current draw patterns, which are consistent with typical values reported by drone manufacturers and practitioners.\footnote{Reference current draw profiles discussed at: \url{https://forum.bitcraze.io/viewtopic.php?t=711}} Additionally, our multi-drone dataset clearly demonstrates the effects of wind and how the recharging station’s availability constraint impacts the delivery time.\footnote{\url{https://bit.ly/3ObYMOm}}

\section{Experiments and Results}\label{experiment}

We conduct a set of experiments to assess the performance of our proposed EPDS framework. Several comparison experiments are designed for the energy consumption prediction module, including different neural network models, feature selections, and sequence length combinations. The RMSE score is used to show how well energy consumption prediction models perform. The proposed composition and optimization approach is then compared to a baseline (A* algorithm), Dijkstra-based, and Bellman-Ford-based approaches. We focus on two main criteria to present the performance of a drone service composition: (1) average delivery time and (2) average execution time. All the tests mentioned above were performed in a Linux environment with an Intel Xeon CPU 2.20 GHz and a Tesla K80 accelerator.

Although the prediction accuracy is generally high and sufficient for scheduling, occasional inaccuracies may occur. When this happens, two situations may arise: (1) a drone arrives earlier than expected and must wait for the recharging station to become available, or (2) a drone arrives later than expected, slightly delaying other drones. In the first case, our system allows the drone to hover. A possible improvement is to delay prediction until the drone has completed part of its flight, thereby increasing accuracy. As a fallback, the system can use A* scheduling without prediction to maintain service continuity. In the second case, while minor delays may occur, the system remains stable and continues to perform better than baseline approaches without prediction.


\subsection{Experimental Setup}
We have created a Python simulation model for drone delivery services to conduct the experiments. Our simulation model is composed of the following modules:
\begin{enumerate*}
    \item request dispatcher,
    \item drone,
    \item node, 
    \item skyway network, 
    \item path planning module,
    \item energy module,
    \item energy consumption prediction module.
\end{enumerate*}
The request dispatcher generates ER from our skyway network's node list and assigns the request to a drone. A drone is a simulation entity with speed, ID, path, current, battery status, and a set of functions. The path planning module creates a path for a drone based on the source and destination nodes, with four different approaches to choose from (Bellman-Ford, Dijkstra, A*, and our proposed heuristic approach). As previously defined, a node is a location in our skyway network with recharging capabilities. The actual node available time and predicted available time are calculated based on the flight information or ECP provided. Based on our real-world dataset, the Energy module updates the battery status of a drone and provides the input sequence for the energy consumption prediction module. The energy consumption prediction module predicts a drone's segment flight energy consumption based on the input sequence.



The topology of our skyway network is built using NetworkX (\cite{hagberg2020networkx}). Our experiments are based on a real-world multi-drone dataset (discussed in Section \ref{data}). Table~\ref{table:attributes} summarizes the dataset's attributes. Table~\ref{table:var} discusses other environmental variables in detail, including the hyperparameters for Bi-LSTM. Since the average segment flight time in our skyway network is around 10 seconds, our goal is to use a short period of time (e.g., 2.5 seconds) to predict the ECP for the whole segment flight. Therefore, the input and prediction lengths are set from 10 to over 100. Here, the timestamp of 100 ms makes the sequence length of 10 equal to 1 second. The hidden size of Bi-LSTM ranges from 32 to 512. We experiment with different learning rates ranging from 0.001 to 0.1 during the model training process. We set the number of drones for all approaches to range between 10 and 50. We collect accurate energy consumption for drones on a small skyway network of 7 nodes, which we then augment to a larger skyway network of 36 nodes. For a path in the larger skyway network, we first use cosine similarity to find a similar segment path in the small skyway network. Then, based on the distance difference, we multiply the energy consumption prediction $ECP$ of the similar path in the small network by a distance ratio. This becomes the $ECP$ for the path in the large skyway network. The speed of the drone ranges from 2 cm/s to 10 cm/s. The Crazyflie drone takes approximately 40 minutes to fully recharge\footnote{https://www.bitcraze.io/support/f-a-q/}. When scaled down to our testbed, the time required to fully recharge a drone ranges between 50 and 150 seconds. For each $ECS$, one source and one destination are randomly generated.

\begin{table}[t]
\centering
\caption{Experimental Variables}
\begin{tabular}{|c|c|}
\hline
\textbf{Variable Name}&{\textbf{Value}} \\
\hline
Input Length &  [10, 125]\\
Prediction Length & [10, 150]\\
Hidden Size & [32, 512]\\
Learning Rate & [0.001, 0.1]\\
Number of Drones & [10, 50]\\
Number of Nodes (recharging stations) & [7, 36]\\
Drone Speed & [2 cm/s, 10 cm/s]\\
Recharging Time& [50 s, 150 s]\\
\hline
\end{tabular}
\label{table:var}
\end{table}

\subsection{Recurrent Neural Network}
For the comparison experiment, we use a vanilla Recurrent Neural Network (RNN) (\cite{sherstinsky2020fundamentals}) with a very similar structure to our energy consumption prediction model. RNN is a memory-based neural network. A vanilla RNN's limitations include one-directional learning, long-range dependency learning, vanishing gradient, and exploding gradient. A one-directional RNN can only learn in one direction (forward). Because of the long-range dependency learning problem, a simple RNN cannot cover a wide range of input. The vanishing gradient means the gradient gets smaller and smaller during back-propagation, resulting in exponential decay. The exploding gradient refers to how rapidly the gradient accumulates, resulting in large updates to the weights in RNN.


\subsection{No-Prediction Approaches}
As previously stated, we compare the performance of our proposed approach to the A*, Dijkstra, and Bellman-Ford approaches. The ECP mechanism is not present in these three approaches. The no-prediction approaches behave similarly to a traditional drone service composition. It only initiates the drone composite service plans and sorts them according to the FCFS policy. No-prediction approaches are typically less efficient when employed in drone delivery services.


\subsection{Results and Discussion}

The proposed approach dynamically optimizes the initial ECS to reduce the waiting time, resulting in a lower delivery time. We first compare our ECP model using different feature selections with RNN for energy consumption prediction. Then, we conduct a comparison experiment for different combinations of sequence lengths. We first compare our initial composition plan with the A*, Dijkstra, and Bellman-Ford approaches to evaluate our proposed approach. Then, we compare the performance of these approaches involving energy consumption prediction and no-prediction. We use (1) RMSE score, (2) execution time, and (3) average delivery time as performance metrics to evaluate the above-mentioned approaches.

\begin{table}[t]
\centering
\caption{Model Comparison and Different Feature Selections}
\begin{tabular}{|c|c|c|}
\hline
\textbf{Model}&{\textbf{Feature Selection}}&{\textbf{RMSE}} \\
\hline
Vanilla RNN & $Vbat$ & 0.1348\\
Bi-LSTM & $Vbat$ & 0.1250\\
Bi-LSTM & All features & 0.1305\\
Bi-LSTM & All features (PCA) & 0.1280\\
\hline
\end{tabular}
\label{table:res1}
\end{table}

\subsubsection{Results for Energy Consumption Prediction}

\begin{itemize}
    \item \textbf{Model Comparison and Different Feature Selections:} Bi-LSTM outperforms vanilla RNN for sequence learning tasks because it overcomes the limitations of unidirectional learning and vanishing gradients. We chose Bi-LSTM based on empirical evidence after comparing it to other models. A drone's battery voltage fluctuation directly reflects the impact of intrinsic and extrinsic constraints. Therefore, learning from the voltage sequence would be better than learning from other feature sequences (e.g., wind speed) for a model. Table~\ref{table:res1} shows the RMSE score of different feature selections with different models. We observe that Bi-LSTM performs better than Vanilla RNN in terms of accuracy. Moreover, using only $Vbat$ as the feature selection shows a better result than other features in our Feature Set. Using PCA slightly improves the accuracy of other features; however, its accuracy is still lower than using only $Vbat$.

 \begin{figure}[t]
     \centering
     \includegraphics[width=0.75\linewidth]{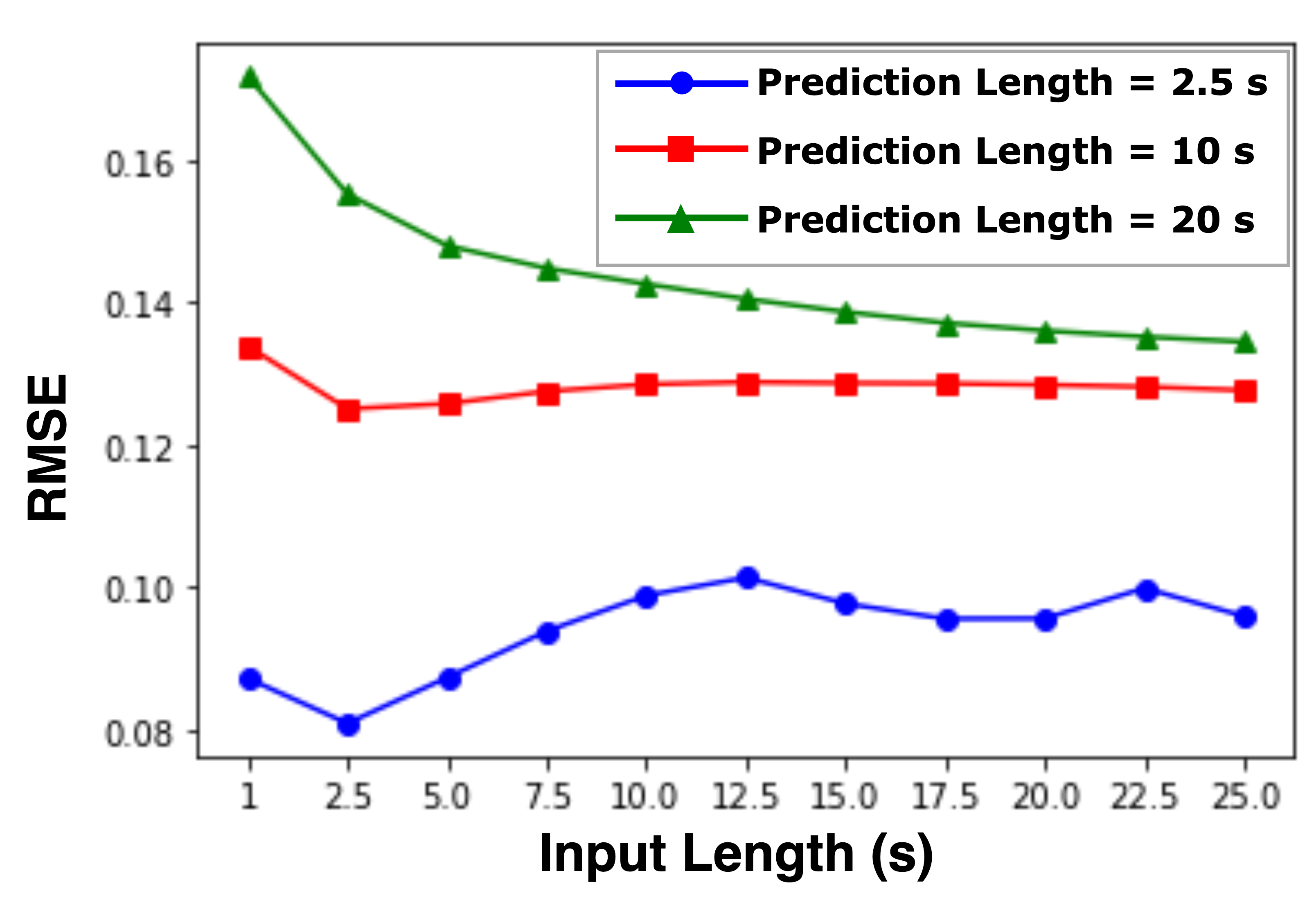}
     \caption{Impact of Sequence Length on ECP}
     \label{fig:window size}
 \end{figure}

\begin{figure}[h]
     \centering
     \includegraphics[width=0.75\linewidth]{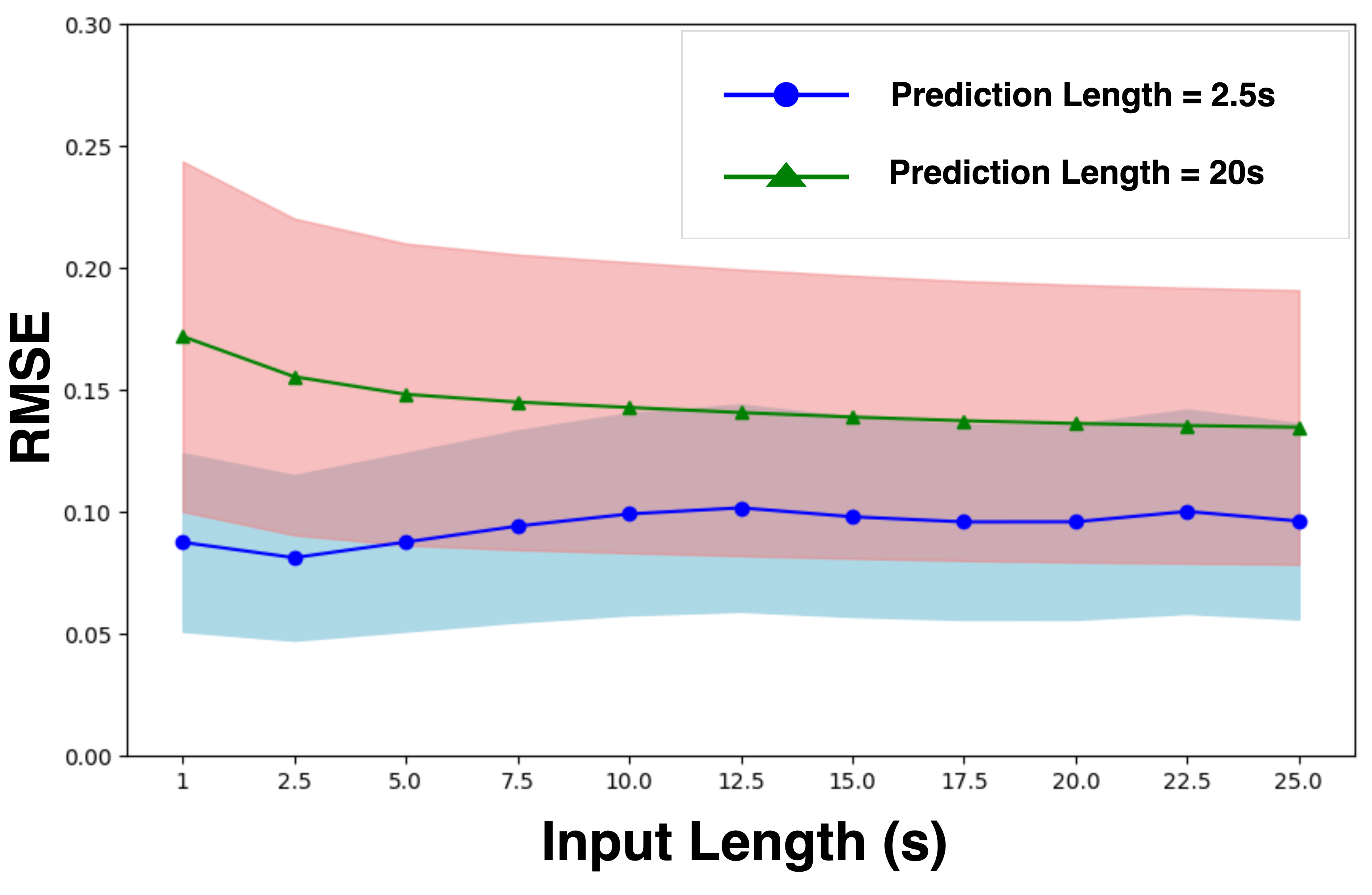}
     \caption{95\% Confidence Intervals for RMSE with Different Prediction Lengths}
     \label{fig:confidence}
 \end{figure}
 
    \item \textbf{Sequence Length: } Selecting the optimal sequence length for input and prediction sequences is essential for predicting energy consumption. In the context of drone delivery services, the ideal combination of sequence length for input and prediction sequences is using a short period of time of in-flight data to predict the ECP for the whole flight segment. Figure \ref{fig:window size} shows the RMSE score of different input and prediction length combinations. Figure \ref{fig:confidence} presents the 95\% confidence intervals for the RMSE scores across varying combinations of input and prediction lengths. The different colors correspond to the confidence intervals for each selected prediction length (i.e., 2.5 s and 20 s), with overlapping regions producing a third blended color. These intervals provide a quantifiable measure of uncertainty around the RMSE values, aiding in the interpretation of the model's performance robustness. We observe that the RMSE score substantially fluctuates when the prediction length is short (2.5 s). However, it becomes relatively smoother as the input length increases. When the prediction length is higher than 2.5 s, the RMSE scores drop rapidly for the smaller input length. However, it stabilizes when the input length increases. A shorter input length is preferred for a longer prediction length for ECP. This is because a short input length helps predict energy consumption and quickly updates the takeoff time for other drones. For our drone model, the ratio of recharging to discharging time is $ratio = T_{recharge} / T_{discharge}$, which is about 7.14 (3000 s / 420 s). In our multi-drone dataset, the flight segment that consumes the most energy is 140 cm long. Based on a drone speed of 6 cm/s, the flight time is approximately 23.3 seconds. To maintain the same ratio (7.14), the recharging time should be approximately 166 seconds; therefore, we set the average recharging time to 150 seconds in our experiment.
\end{itemize}

\begin{figure*}[t]
\minipage{0.49\textwidth}
\centering
  \includegraphics[width=\linewidth]{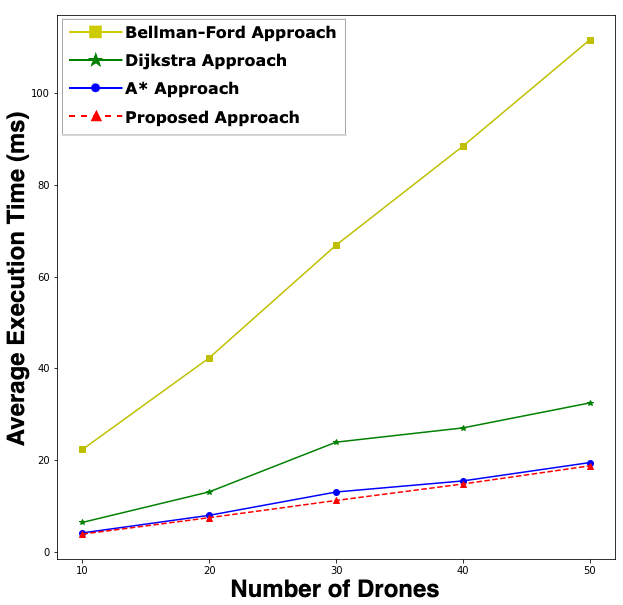}
    \caption{Average Execution Time}
    \label{fig:exe}
\endminipage\hfill
\minipage{0.49\textwidth}%
\centering
  \includegraphics[width=\linewidth]{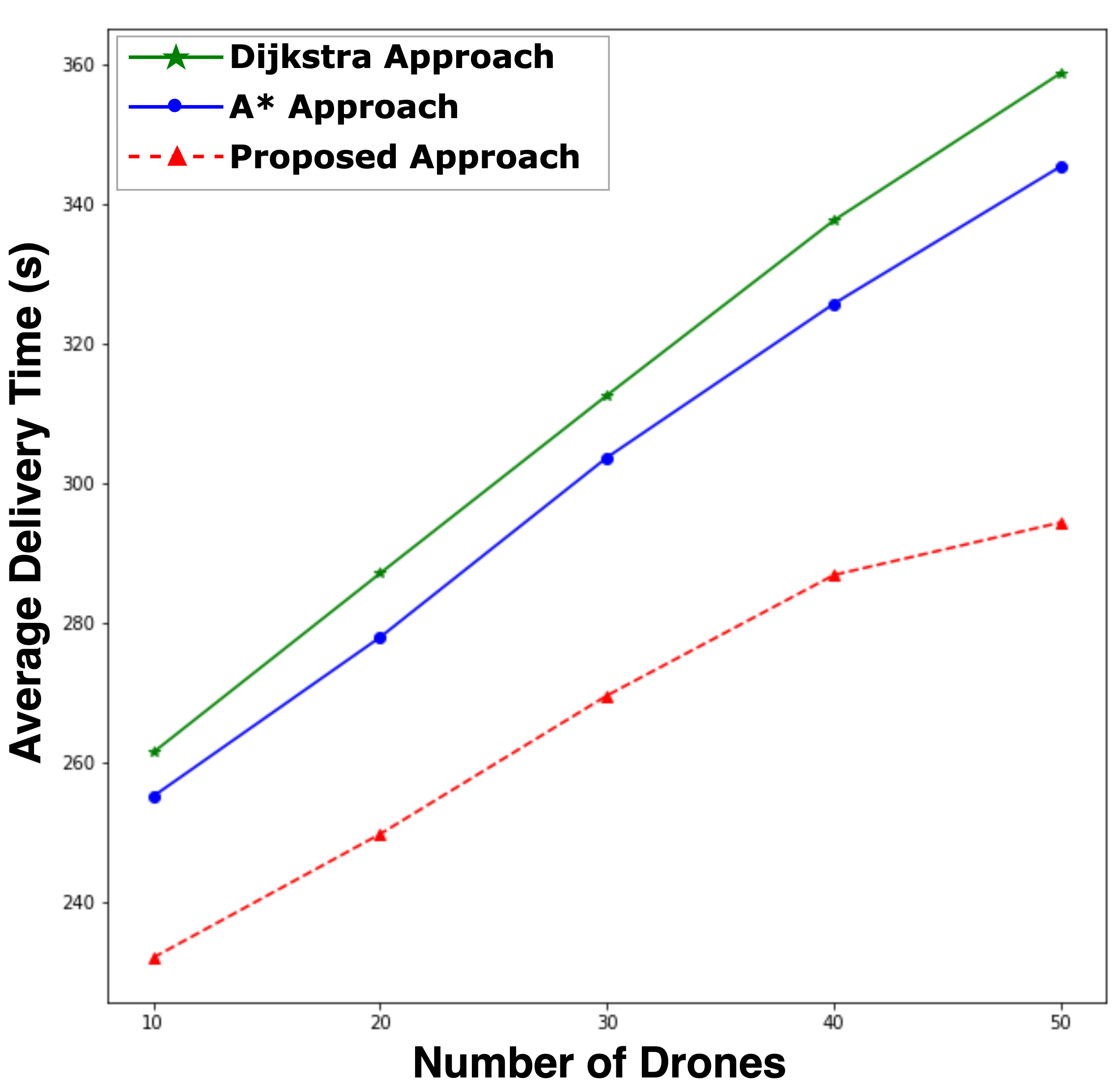}
    \caption{Average Delivery Time}
    \label{fig:overall}
\endminipage
\end{figure*}

\begin{figure*}[t]
\centering
\minipage{0.49\textwidth}
\centering
  \includegraphics[width=\linewidth]{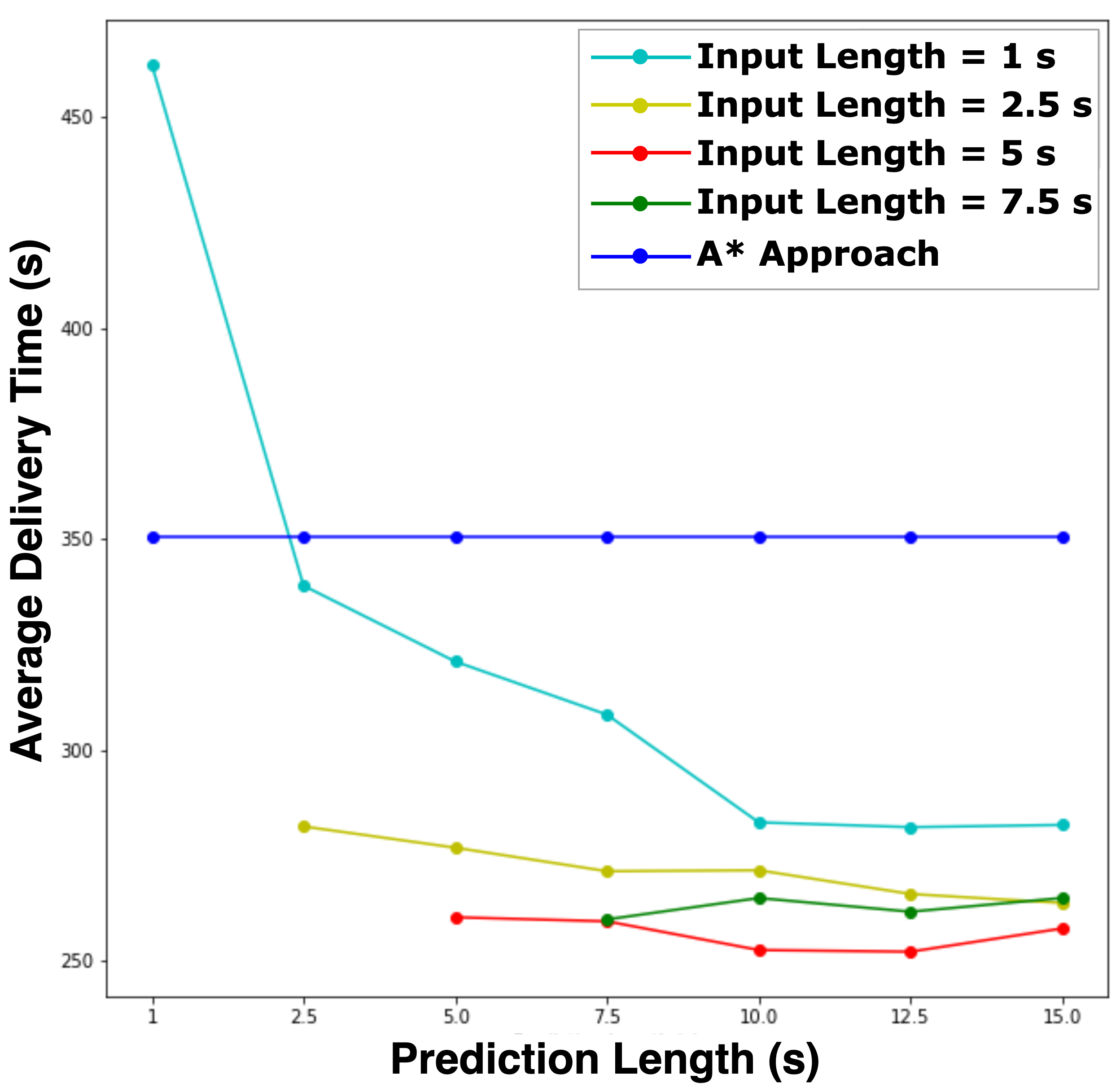}
    \caption{Sequence Length Impact on Average Delivery Time}
    \label{fig:overall-window size}
\endminipage\hfill
\minipage{0.49\textwidth}%
\centering
  \includegraphics[width=\linewidth]{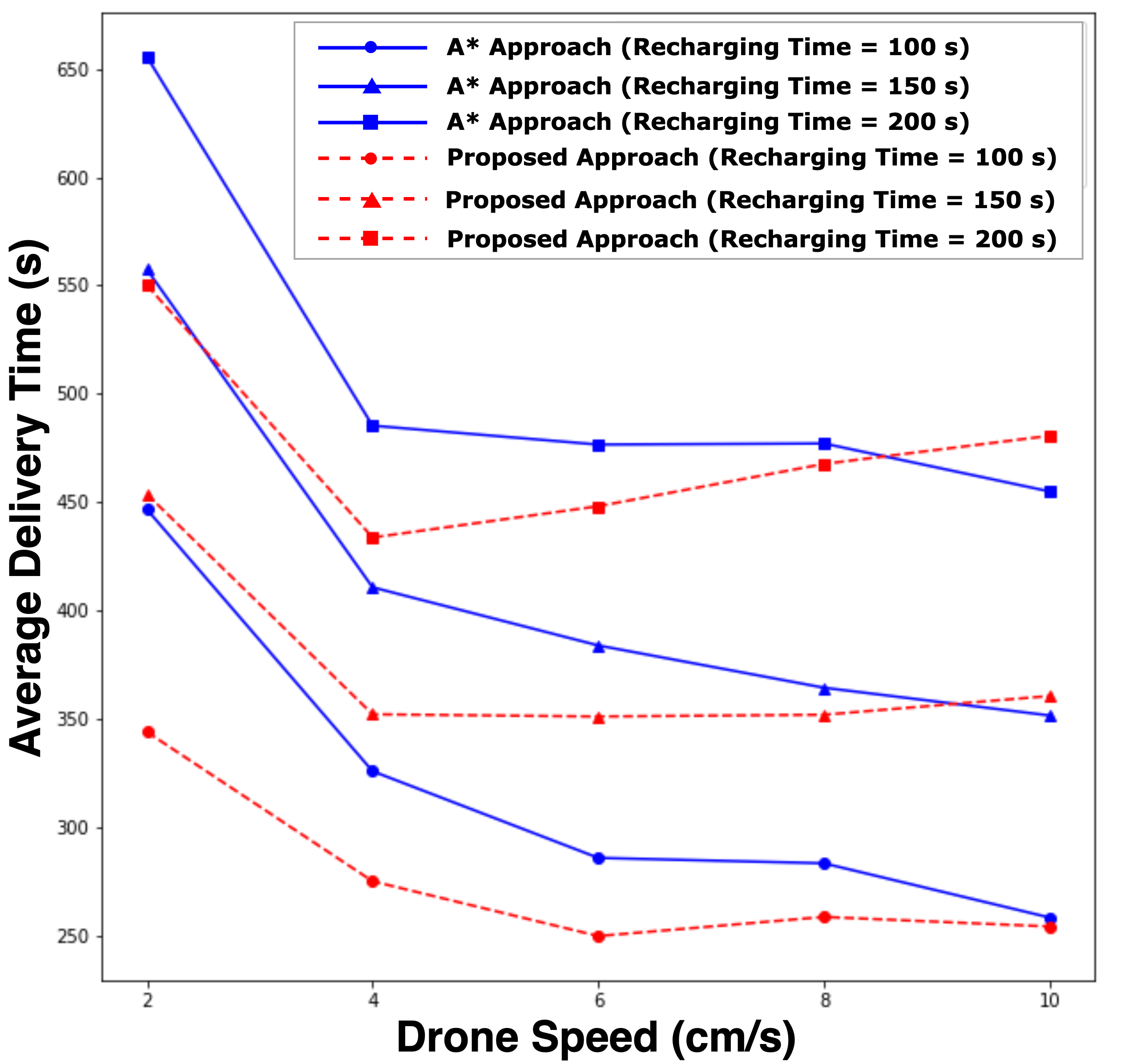}
    \caption{Drone Speed and Recharging Time Impact}
    \label{fig:speed}
\endminipage
\end{figure*}

\subsubsection{Results for Optimization of Service Composition}
\begin{itemize}
    \item \textbf{Average Execution Time:} The time complexities of Bellman-Ford, Dijkstra, A*, and our proposed approaches are discussed in section \ref{timecomplxitysection}. The time complexity of Bellman-Ford is relatively high. Figure \ref{fig:exe} presents the average execution times for these approaches. As expected, the execution time of Bellman-Ford is higher than that of other approaches. We observe that the Bellman-Ford's execution time increases rapidly as the number of drones increases. This is because there are more possible composition plans. Dijkstra's execution time increases significantly slower than the Bellman-Ford approach. The results demonstrate that our proposed approach computes the solutions significantly faster than all other approaches.
    
    \item \textbf{Average Delivery Time:} 
    In our experiments, the Dijkstra and Bellman-Ford algorithms produce identical composition plans in the skyway network. This is expected in our setting, where all edge weights are non-negative and the graph structure is static. Since both algorithms compute shortest paths under these conditions, including Bellman-Ford in the comparisons would be redundant. To maintain clarity, we therefore present results only for Dijkstra and A*. Our proposed approach reduces drones' waiting time by predicting energy consumption when an RSC occurs. Figure \ref{fig:overall} shows the average delivery times of a single drone in the drone services composition for Dijkstra, A*, and our proposed approaches. The results clearly illustrate that our proposed approach outperforms the other two approaches.
    
    \item \textbf{Impact of Sequence Length:} Sequence length for input and prediction sequence is important to energy consumption prediction and the average delivery time. Our proposed approach uses two strategies for the ECP task: chained prediction and clipping. Chained prediction is applied when the prediction length is not long enough to cover the whole segment flight. Clipping refers to the process of shortening the prediction sequence from the beginning to the length of the segment flight if the prediction length is longer than the segment flight. Figure \ref{fig:overall-window size} shows how the selection of sequence lengths impacts the average delivery time compared to the A* approach. We observe that the average delivery time becomes extremely high when the input and prediction lengths are short. This increase in delivery time is due to the exponential increase in prediction uncertainties over prediction. As the prediction length increases, all curves tend to be stable. We observe that using an input length ranging from 2.5 s to 7.5 s has promising performance for a long prediction length (10 s to 15 s).
    
    \item \textbf{Impact of Drone Speed and Recharging Time:} Our proposed approach reduces the average delivery time by reducing the waiting time using energy consumption prediction. The waiting time is directly related to the drone's speed and recharging time. A slower drone speed has the same effect as a long-distance flight, while a shorter recharging time to achieve full battery capacity has the same effect as a faster recharging rate. Figure \ref{fig:speed} shows how drone speed and recharging time impact the average delivery time of our proposed approach compared to the A* approach. We observe that our proposed approach is superior to A* for long-distance services (i.e., slow drone speed). As the recharging rate goes faster, our proposed approach performs better than a lower recharging rate.

    \item \textbf{Impact of Varying Network Sizes} Our proposed approach shows better performance on large-scale networks and small-scale networks in terms of average delivery time and average execution time. We use different numbers of nodes to form various sizes of fully connected networks with a fixed number of drones at 10. The superiority of our approach becomes more evident as the network size increases. Figure \ref{fig:node_exe} and Figure \ref{fig:node_dt} show how network size impacts the average delivery time and average execution time of our approach compared to other baseline approaches, respectively.

\begin{figure*}[t]
\centering
\minipage{0.49\textwidth}
\centering 
  \includegraphics[width=\linewidth]{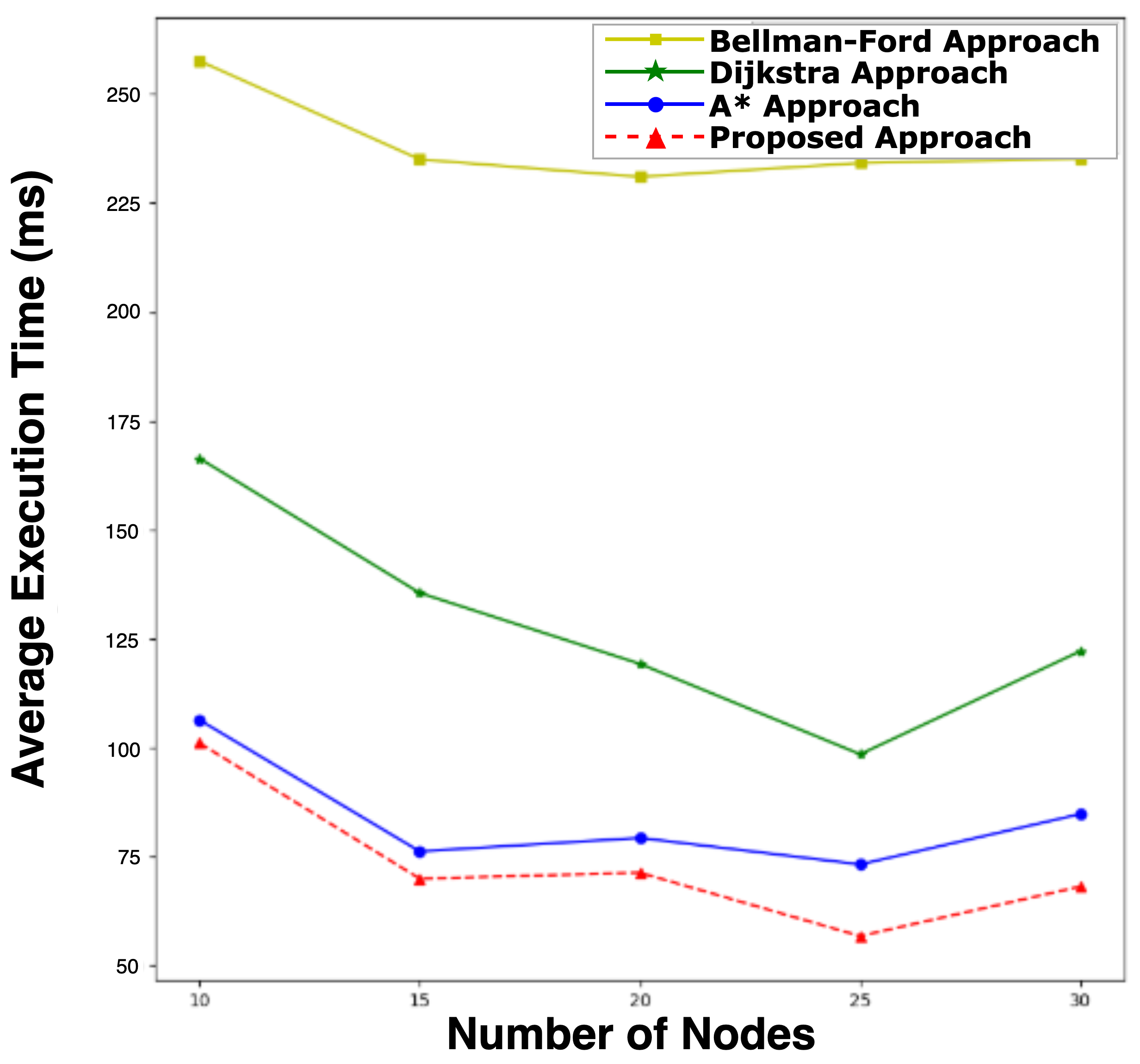}
    \caption{Network Size Impact on Average Execution Time}
    \label{fig:node_exe}
\endminipage\hfill
\minipage{0.49\textwidth}%
\centering
  \includegraphics[width=\linewidth]{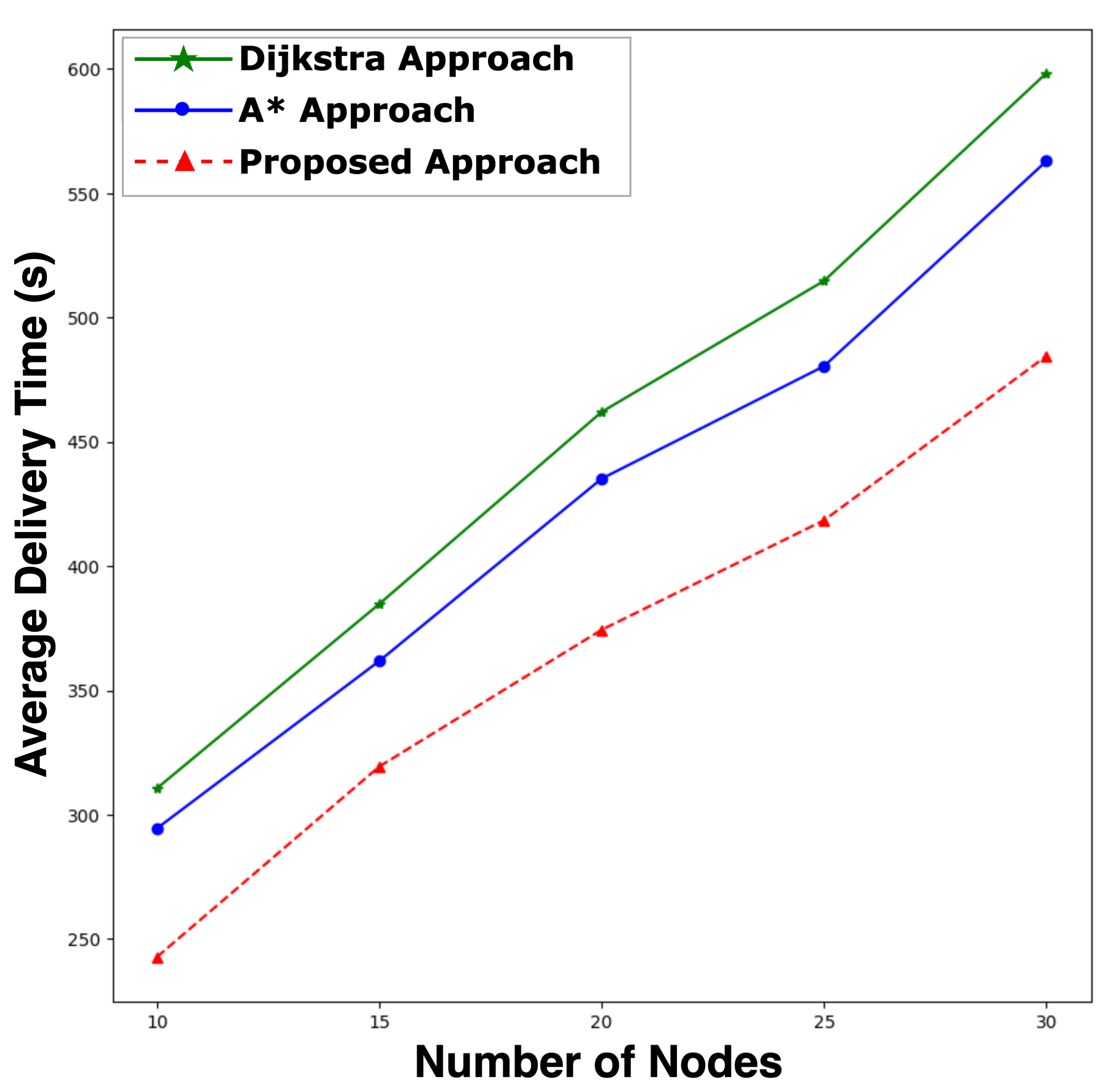}
    \caption{Network Size Impact on Average Delivery Time}
    \label{fig:node_dt}
\endminipage
\end{figure*}
\end{itemize}


\section{Conclusion}

We presented an Energy-Predictive Drone Service (EPDS) composition framework that addresses efficient drone delivery in skyway networks. The framework comprises three key modules: an initial composition module, an energy consumption prediction (ECP) module, and an EPDS optimization module. The initial composition module generates an energy- and time-efficient drone service composition plan for a given customer request. The ECP module is adaptable to varying flight lengths and provides energy consumption predictions for each segment. The service optimization algorithm then dynamically refines the initial plan using the ECP module’s predictions.

We conducted extensive experiments comparing our framework against another deep learning model and multiple baseline approaches that do not use prediction. The results demonstrate that our framework is runtime-efficient and significantly reduces the average delivery time by approximately 14\% compared to shortest-path algorithms such as Dijkstra's and A*. Furthermore, in scenarios with long-distance flights and fast recharging techniques, our framework outperformed the A* approach by approximately 16\% in terms of average delivery time.

In this work, we treat payload weight as a fixed constant to reflect a single-package delivery scenario and the direct impact of payload weight on a drone's voltage during flight. We plan to investigate energy consumption prediction under varying payload types and weights, as well as their interaction with dynamic wind conditions, in future work to further enhance the real-world applicability of our approach.

\section*{Acknowledgement}

This research was partly made possible by LE220100078 and DP220101823 grants from the Australian Research Council. The statements made herein are solely the responsibility of the authors.

 \bibliographystyle{elsarticle-num-names} 
 \bibliography{ref}





\end{document}